%% file: main.tex
\documentclass[]{fairmeta}

\input{math_commands.tex}

\usepackage{subcaption}
\usepackage{wrapfig}
\usepackage[most]{tcolorbox}
\usepackage[most]{tcolorbox}
\usepackage{xcolor}
\usepackage{hyperref}
\usepackage{wrapfig}

\definecolor{softgreen}{RGB}{110, 160, 120}
\usepackage{svg}
\usepackage[most]{tcolorbox}

\newtcolorbox{takeawaybox_basemodel}[1]{
    colback=orange!5!white,
    colframe=black,
    arc=5pt,
    outer arc=5pt,
    boxrule=0.8pt,
    left=5pt,
    right=5pt,
    top=4pt,
    bottom=4pt,
    fontupper=\small,
    enhanced,
    before upper={\textbf{#1 }} 
}

\newtcolorbox{promptbox}[1][]{
    colback=gray!5,
    colframe=gray!50,
    fonttitle=\bfseries\small,
    title=#1,
    breakable,
    left=4pt, right=4pt, top=4pt, bottom=4pt,
    fontupper=\small\ttfamily,
}

\usepackage{caption}
\usepackage{wrapfig}
\usepackage{booktabs}
\definecolor{earlyblue}{HTML}{88A2F1}
\definecolor{midgrey}{HTML}{fadcb4}
\definecolor{latered}{HTML}{EE9C88}
\definecolor{highlightgreen}{HTML}{80c66d}
\definecolor{highlightpurple}{HTML}{9b6d97}
\def\thickhline{\noalign{\hrule height.8pt}}
\usepackage{arydshln}
\usepackage{xstring}
\newcommand{\deltaval}[1]{%
  \IfBeginWith{#1}{+}{%
    {\textcolor{highlightgreen}{\textit{(#1)}}}%
  }{%
    \IfBeginWith{#1}{-}{%
      {\textcolor{highlightpurple}{\textit{(#1)}}}%
    }{%
      {\textit{(#1)}}%
    }%
  }%
}
\usepackage{pifont}
\usepackage{epigraph}
\usepackage{hyperref}

\usepackage[export]{adjustbox}

\newcommand{\benchmarkmini}{\text{Auto-ClawEval-Mini}\xspace}
\newcommand{\ours}{\text{ClawEnvKit}\xspace}
\newcommand{\benchmark}{\text{Auto-ClawEval}\xspace}

\title{\ours : Automatic Environment Generation for Claw-Like Agents}
\author[1]{Xirui Li}
\author[1,5]{Ming Li}
\author[2,3]{Ion Stoica}
\author[2,4]{Cho-Jui Hsieh}
\author[5]{Tianyi Zhou}

\renewcommand\affiliation[2][]{%
  \addtolist[#1]{#2}{\affiliationlist}{\affiliationformat}{\\}%
}

\affiliation[1]{University of Maryland}
\affiliation[2]{Arena}
\affiliation[3]{University of California, Berkley}
\affiliation[4]{University of California, Los Angeles}
\affiliation[5]{Mohamed bin Zayed University of Artificial Intelligence}
\abstract{
\input{tex/0_abstract}
}

\date{\today}
\authoremails{\email{xiruili@umd.edu}, 
\email{tianyi.zhou@mbzuai.ac.ae}}

\metadata[Project Page]{\url{https://github.com/xirui-li/ClawEnvKit}\\}

\begin{document}

\maketitle

\input{tex/1_introduction}

\input{tex/2_background}

\input{tex/3_problem_formulation}

\input{tex/4_system}

\input{tex/5_experiments}
\input{tex/6_conclusion}

\clearpage
\newpage
\bibliographystyle{assets/plainnat}
\bibliography{main}

\clearpage
\tableofcontents
\newpage
\beginappendix

\input{tex/7_appendix}

\end{document}

%% file: math_commands.tex
\usepackage{amsmath,amsfonts,bm}
\usepackage{xspace}
\usepackage{hyperref}
\usepackage{url}
\usepackage{subcaption}

\usepackage{colortbl}
\usepackage{xcolor}

\usepackage{graphicx}
\usepackage{algorithmic}
\usepackage{algorithm}
\usepackage{float}

\usepackage{booktabs}

\usepackage{enumitem}

\usepackage{amsthm}

\theoremstyle{plain}
\newtheorem{theorem}{Theorem}[section]

\theoremstyle{definition}
\newtheorem{definition}[theorem]{Definition}

\theoremstyle{remark}

\def\eqref#1{equation~\ref{#1}}

\def\1{\bm{1}}

\DeclareMathAlphabet{\mathsfit}{\encodingdefault}{\sfdefault}{m}{sl}
\SetMathAlphabet{\mathsfit}{bold}{\encodingdefault}{\sfdefault}{bx}{n}

%% file: tex/0_abstract.tex
Constructing environments for training and evaluating claw-like agents remains a manual, human-intensive process that does not scale.
We argue that what is needed is not just a dataset, but an \emph{automated pipeline} capable of generating diverse, verified environments on demand. 
To this end, 
% we formalize an environment as a three-tuple $E = (P, M, C)$: a task specification $P$, a tool interface $M$, and a scoring configuration $C$.
we introduce \textbf{ClawEnvKit}, an autonomous generation pipeline that instantiates this formalism from natural
language descriptions. The pipeline comprises three modules: (1) a \emph{parser} that extracts structured generation parameters from natural language input; (2) a \emph{generator} that produces the task specification, tool interface, and scoring configuration; and (3) a \emph{validator} that enforces feasibility, diversity, structural validity, and internal consistency across the generated environments. Using ClawEnvKit, we construct \textbf{\benchmark}, the first large-scale benchmark for claw-like agents, comprising 1{,}040 environments across 24 categories. 
Empirically, \benchmark matches or exceeds human-curated environments on coherence and clarity at 13{,}800$\times$ lower cost.
Evaluated across 4 model families and 8 agent harness frameworks, we find that harness engineering boosts performance by up to 15.7 percentage points over a bare ReAct baseline, completion remains the primary axis of variation with no model saturating the benchmark, and automated generation enables evaluation at a scale previously infeasible.
Beyond static benchmarking, \ours enables \textbf{live evaluation}: users describe a desired capability in natural language and obtain a verified environment on demand, turning evaluation into a continuous, user-driven process.
The same mechanism serves as an on-demand training environment generator, producing task distributions that adapt to an agent's current weaknesses rather than being bounded by existing user logs.

%% file: tex/1_introduction.tex
\section{Introduction}
\label{sec:introduction}

Large language model (LLM) agents are increasingly being deployed in real-world environments to autonomously handle complex, multi-step tasks~\citep{yao2023reactsynergizingreasoningacting,
shinn2023reflexionlanguageagentsverbal}. By equipping LLM agents with harness~\citep{openai2026harness, lee2026metaharnessendtoendoptimizationmodel, anthropic2025harness, boluk2026harness, boeckeler2026harness}, they extend beyond static text generation to actively interact with digital ecosystems, including file systems, web services, and application programming interfaces (APIs). Exemplified by \emph{claw-like agents}, such as \texttt{OpenClaw}~\citep{openclaw2025}, \texttt{NanoClaw}~\citep{nanoclaw2026}, and \texttt{IronClaw}~\citep{ironclaw2026}, the rapid proliferation of such systems signals a broader paradigm shift from LLMs as passive language interfaces to LLM-driven agents as autonomous actors embedded in real-world scenarios.

To investigate and improve claw-like agents in real-world scenarios, researchers~\citep{xia2026metaclaw, wang2026openclawrl, claw-eval2026, ji2026clawarenabenchmarkingaiagents} construct \emph{environments} for training and evaluation that specify (1) the executable scenarios defining what an agent must do, (2) the tools it can use, and (3) how its actions are verified. OpenClaw-RL~\citep{wang2026openclawrl} and MetaClaw~\citep{xia2026metaclaw} improve agent capabilities via reinforcement learning on trajectories collected from real user environments, while Claw-Eval~\citep{claw-eval2026} and SkillsBench~\citep{li2026skillsbench} provide human-curated environments for evaluating such systems. However, both directions face fundamental limitations: training is constrained to whatever tasks users happen to perform, and benchmarks require hundreds of person-hours to construct yet become static once released. This shared bottleneck, the cost and rigidity of manual environment construction, prevents training and evaluation from scaling alongside rapidly advancing agent capabilities.

We present \ours, a scalable framework that automates agent environment generation for claw-like agents on demand. Given a natural language specification, \ours produces verified agent environments in which agents interact with mock services and are graded automatically, reducing the cost of environment construction from hours of human labor to minutes of automation. The pipeline comprises three modules: (1) a \textbf{Parser} that converts natural language into structured specifications, (2) a \textbf{Generator} that instantiates task environments, and (3) a \textbf{Validator} that enforces structural and semantic correctness. In each generated environment, the agent runs in an isolated sandbox that supports the full family of claw-like agent harnesses and models, supporting long-horizon tasks without cross-task interference. Empirically, we show that automatically generated environments match or exceed human-curated ones on all quality dimensions while reducing construction cost and time.

Building on \ours, we automatically construct two benchmarks based on services from Claw-Eval. \textbf{\benchmark} contains 1{,}040 environments spanning 24 semantic categories for the first-ever large-scale cross-harness evaluation, and \textbf{\benchmarkmini} is a compact 104-task version paired one-to-one with Claw-Eval for direct quality comparison. Experiments across 8 agent harness frameworks and 4 model families reveal that harness engineering is a significant performance booster: all structured harnesses outperform the ReAct baseline by up to 15.7 percentage points, confirming that \benchmark is not saturated by current frontier models. Scores on the full \benchmark and the compact \benchmarkmini differ by less than 2\%, validating that automated generation can reliably scale benchmark size without sacrificing evaluation quality. 

Beyond static benchmarking, \ours enables \textbf{live evaluation}: users describe a desired capability in natural language and obtain a verified environment on demand, turning evaluation into a continuous, user-driven process that keeps pace with emerging tasks and long-tail domains. The same mechanism doubles as an on-demand training environment generator, producing task distributions that adapt to an agent's current weaknesses rather than being bounded by existing user logs.
\begin{figure}[t]
    \centering
    \includegraphics[width=0.8\textwidth]{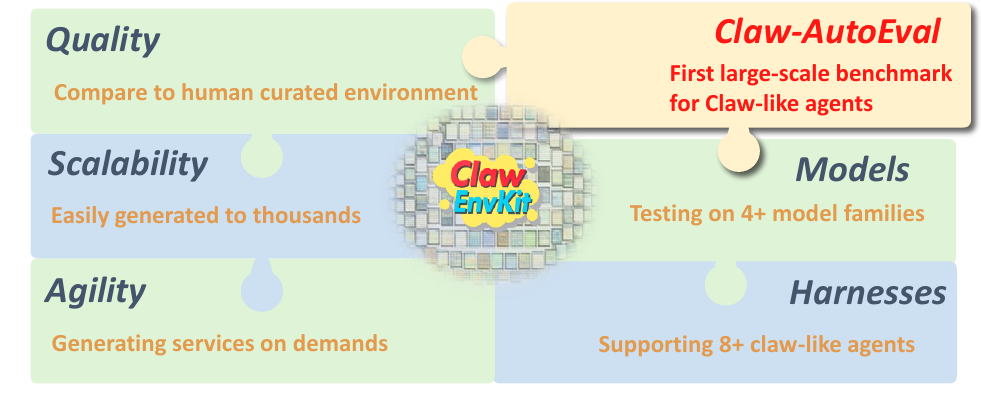}
    \caption{
    \textbf{\ours at a glance.} \ours provides three key properties (left): \emph{quality} comparable to human-curated benchmarks, \emph{scalability}
    to an unlimited number of environments, and\emph{adaptability} through on-demand curation. The framework ships with supports 4+ model families, and integrates with 8+ claw-based agent harnesses out of the box (right).
    }
    \label{fig:clawenvkit_pipeline}
\end{figure}

Our main contributions are:
\begin{enumerate}
  \item \textbf{\ours}, a scalable framework for automated agent environment generation that separates declarative specification from deterministic verification, runs each task in an isolated sandbox preserving agent-native workflows, and supports the full family of claw-based agents across multiple backbone models.

  \item \textbf{The first large-scale benchmark, \benchmark,} spanning 24 domains, evaluated across claw-based agents and backbone models, serving as the first large-scale, cross-harness, cross-backbone benchmark in the claw ecosystem.
 
  \item \textbf{Live evaluation}, where end users generate bespoke evaluation cases on demand through natural language, is demonstrated by our \ours framework.
\end{enumerate}

%% file: tex/2_background.tex
\section{Relate Work}
\label{sec:related_work}

\subsection{Scaling up Environment Generation}
\label{sec:related-scaling}
 
Constructing agent environments has been a manual, labor-intensive process.
AgentBench~\citep{liu2023agentbench} provides hand-crafted interactive 
environments for multi-turn LLM evaluation, finding a large capability gap 
between commercial and open-source models.
GUI benchmarks~\citep{sun2022meta, lu2024weblinx, OSWorld, 
chen2025guiworldvideobenchmarkdataset} build high-fidelity web or GUI 
environments for functional task execution but require significant engineering 
effort per domain. Web agent frameworks~\citep{zhou2023webarena, 
workarena2024, dechezelles2025browsergymecosystemwebagent,koh2024visualwebarenaevaluatingmultimodalagents} pursue reproducibility through self-hosted applications and Gym-style evaluation, yet static benchmarks degrade as live interfaces evolve, motivating online evaluation 
methods~\citep{pan2024webcanvasbenchmarkingwebagents, yoran2024assistantbenchwebagentssolve} and continuously 
updated task sets~\citep{zhang2025swebenchgoeslive}.
On the infrastructure side, sandboxed agent  platforms~\citep{wang2025openhands_sdk} and environment configuration benchmarks~\citep{eliseeva2025envbenchbenchmarkautomatedenvironment} address execution safety and dependency resolution, but each remains purpose-built for a specific domain.

Recent work has begun to address this scalability bottleneck through automatic environment synthesis. AgentStudio~\citep{zheng2024agentstudio} provides a toolkit for building general virtual agents with tools for creating online benchmark tasks across GUI and API action spaces. SWE-smith~\citep{yang2025swesmithscalingdatasoftware} automatically constructs software engineering tasks from GitHub repositories by seeding bugs and filtering with test execution. R2E-Gym~\citep{jain2025r2egymproceduralenvironmentshybrid} uses a data curation pipeline to synthesize executable coding environments. RandomWorld~\citep{sullivan-etal-2025-procedural} procedurally generates tool-use environments for API-calling agents.
Agent World Model~\citep{wang2026agent} synthesizes executable tool-use environments at scale by decomposing generation into a stateful backend, a tools interface layer, and task-specific success criteria. 
Endless Terminal~\citep{gandhi2026endlessterminalsscalingrl} provides a pipeline that procedurally generates terminal-use tasks without human annotation. Our work is the first of the kind to provide scalable environment for claw-like agents that we discuss as follows.

\begin{table*}[htbp]
\centering
\small
\caption{
\textbf{Comparison of environments that evaluate claw-like agents.} \benchmark is the only framework that combines auto-generated tasks, universal verification, continuous scoring, safety gates, robustness testing, and support for the full family of claw-like agents. Claw-Eval is a growing benchmark, we use the version snapshot on \texttt{2026-04-01}.}
\label{tab:comparison}
\resizebox{\textwidth}{!}{
\begin{tabular}{p{5.0 cm} c c c c c c c c}
\toprule
\textbf{Evironments}
& \textbf{Tasks}
& \textbf{Source}
& \textbf{Grading}
& \textbf{Generalizability}
& \textbf{Safety Eval}
& \textbf{Robustness Eval}
& \textbf{Harness support} \\
\midrule

ClawArena~\citep{ji2026clawarenabenchmarkingaiagents}
& 64
& Human
& Binary
& \ding{55}
& \ding{55}
& \ding{55}
& \ding{51} \\

ClawsBench~\citep{li2026clawsbenchevaluatingcapabilitysafety}
& --
& Human 
& Rubric
& \ding{55}
& \ding{51}
& \ding{51}
& \ding{55} \\

% ClawArena~\citep{ji2026clawarenabenchmarkingaiagents}
% & 64
% & Human
% & Multiple Choice
% & Binary
% & \ding{55}
% & \ding{55}
% & \ding{51} \\

% & \ding{55} \\
SkillsBench~\citep{li2026skillsbench}
& 84
& Human
& Binary
& \ding{55}
& \ding{55}
& \ding{55}
& \ding{55} \\

Claw-Eval$^{*}$~\citep{claw-eval2026}
& 104
& Human
& Rubric
& \ding{55}
& \ding{51}
& \ding{51}
& \ding{55} \\

\midrule
\rowcolor{blue!8}
\textbf{\benchmark (ours)}
& \textbf{1{,}040}  
& \textbf{Auto-generated}
& \textbf{Rubric}
& \ding{51}
& \ding{51}
& \ding{51}

& \ding{51} \\
\bottomrule
\end{tabular}
}
\end{table*}

\subsection{Claw-like Agents}
\label{sec:related-claw}
The claw-like agent ecosystem~\citep{openclaw2025} provides a family of open-source CLI agent platforms (\texttt{OpenClaw}~\citep{openclaw2025}, \texttt{NanoClaw}~\citep{nanoclaw2026}, \texttt{IronClaw}~\citep{ironclaw2026}, and others) that interact with external services through
native tool calls and support continue-learning~\citep{wang2024comprehensive} by modifying skills markdown. 

On the training side, OpenClaw-RL~\citep{wang2026openclawrl} and MetaClaw~\citep{xia2026metaclaw} scale agent training by collecting trajectories from real user interactions, but remain limited by the diversity and volume of available usage data.
Recent benchmarks such as ClawArena~\cite{ji2026clawarenabenchmarkingaiagents}, ClawsBench~\citep{li2026clawsbenchevaluatingcapabilitysafety}, Claw-Eval~\citep{claw-eval2026}, and SkillsBench~\citep{li2026skillsbench} evaluate agent capabilities across dynamic information environments, realistic productivity workflows, and structured API tasks; however, they all rely on fixed, human-authored task distributions, limiting scalability, diversity, and coverage of real-world scenarios.
\ours addresses these limitations as a scalable source of environments for both training and evaluation: it synthesizes diverse environments on demand, without requiring
existing user traffic or manual task authoring.  With \ours, we obtain the first large-scale benchmark (\benchmark) for claw-like agents. Table~\ref{tab:comparison} demonstrate a direct comparison with latest benchmarks.

%% file: tex/3_problem_formulation.tex
\section{Formalizing Environments for Claw-like Agents}
\label{sec:formulation}

\begin{figure}[t]
    \centering
    \includegraphics[width=\textwidth]{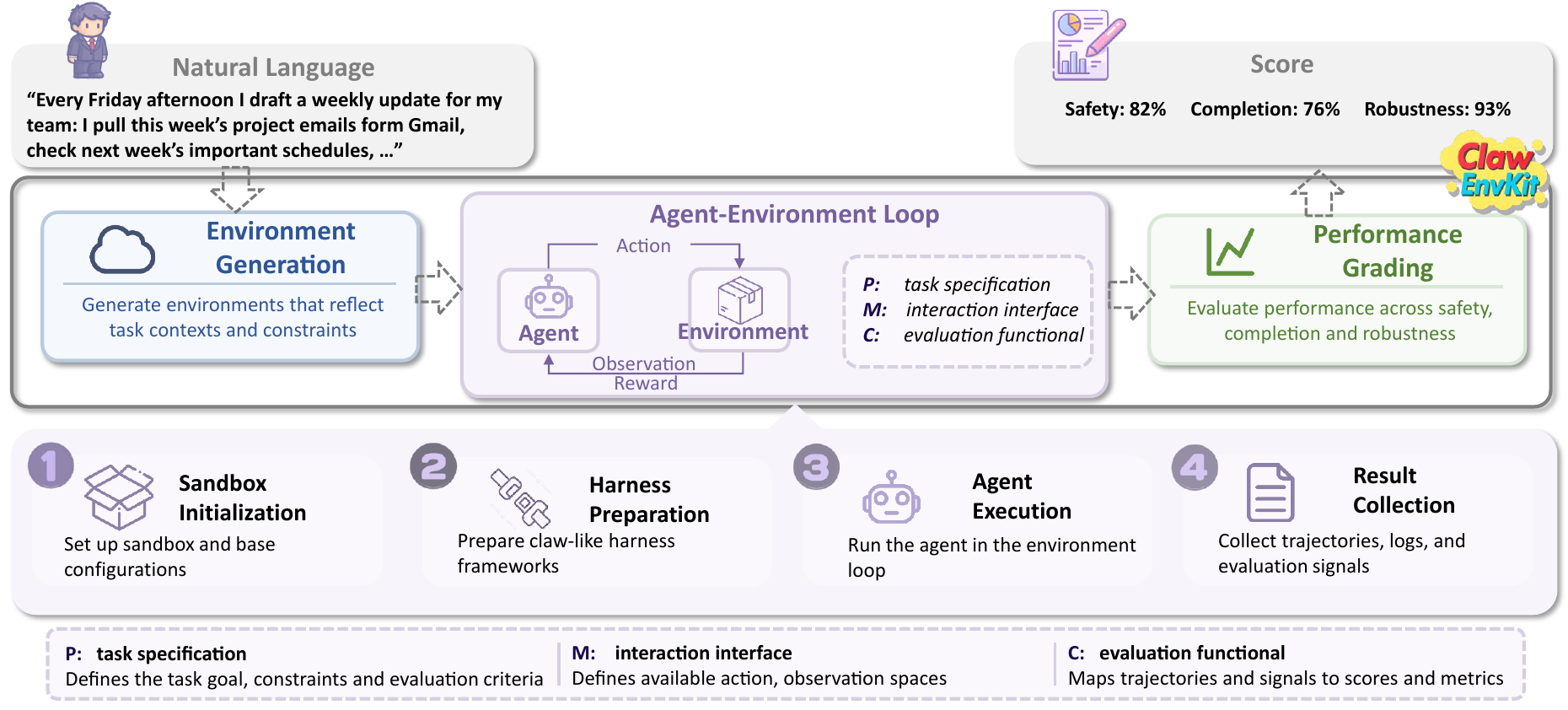}
    \caption{
        \textbf{Overview of the \ours pipeline.} Given a natural language specification (upper left), the \textbf{Environment Generation} module produces a set of $N$ task environments $E = (P, M, C)$, each comprising a task specification $P$, an interaction interface $M$, and an evaluation functional $C$.
        Each environment is then executed through four sequential steps: (1) Sandbox Initialization, (2) Harness Preparation, (3) Agent Execution, and (4) Result Collection. At the end, the \textbf{Performance Grading} module scores the agent trajectory along three dimensions: Safety, Completion, and Robustness (upper right).
    }
    \label{fig:clawenvkit_pipeline}
\end{figure}

Classical environments in reinforcement learning are modeled as Markov Decision Processes with an explicit, enumerable
state space $\mathcal{S}$, a formalism well-suited to bounded domains such as game simulators or robot controllers~\citep{sutton1998reinforcement}. Modern agent settings break this assumption: an agent that reads emails, calls APIs, and reasons over multi-turn conversation histories operates over a state space that is effectively infinite, driven by unbounded natural language context, tool outputs, and interaction history. Yet the \emph{implementation} of such an environment is finite: in our setting, the environment state reduces to the contents of a small number of in-memory mock service databases, fully determined by the fixture data loaded at startup.
This asymmetry, infinite from the agent's perspective, finite from the implementer's, suggests a different representational strategy: rather than specifying the state space, we specify \emph{what the agent must do}~($P$), \emph{what it can do}~($M$), and \emph{how it is evaluated}~($C$).
This declarative separation is what makes automated generation tractable: an LLM can produce a valid $(P, M, C)$
triple without ever reasoning about state transitions, whereas generating a correct state-based grader requires understanding the full execution semantics of the environment.

\begin{definition}[Environment]
\label{def:env}
An \textbf{environment} is a three-tuple
$E = (P,\, M,\, C)$, where:
\begin{itemize}
  \item $P \in \mathcal{L}$ is a \textbf{task specification} in natural language.
  \item $M = (\mathcal{T},\, \mathcal{O})$ is the \textbf{interaction interface}: $\mathcal{T}$ is a set of callable tools and $\mathcal{O}$ is the audit log recording every tool call, its parameters, and its server-side outcome.
  \item $C = \{(c_i,\, w_i)\}$ is the \textbf{evaluation
    functional}, where $\Sigma$ denotes the space of agent
    trajectories and each
    $c_i : \Sigma \times \mathcal{O} \to [0,1]$
    evaluates a property of the agent's trajectory $\sigma
    \in \Sigma$ against the audit log, with
    $\sum_i w_i = 1$.
\end{itemize}
The \textbf{score} of a trajectory $\sigma$ on environment
$E$ is:
\begin{equation}
  R(\sigma, E) \;=\; \sum_i w_i \cdot c_i(\sigma,
  \mathcal{O}).
  \label{eq:score}
\end{equation}
\end{definition}

%% file: tex/4_system.tex
\section{\ours: A Scalable Framework for Automated Environment Generation}
\label{sec:system}

\begin{figure}[t]
    \centering
    \includegraphics[width=\textwidth]{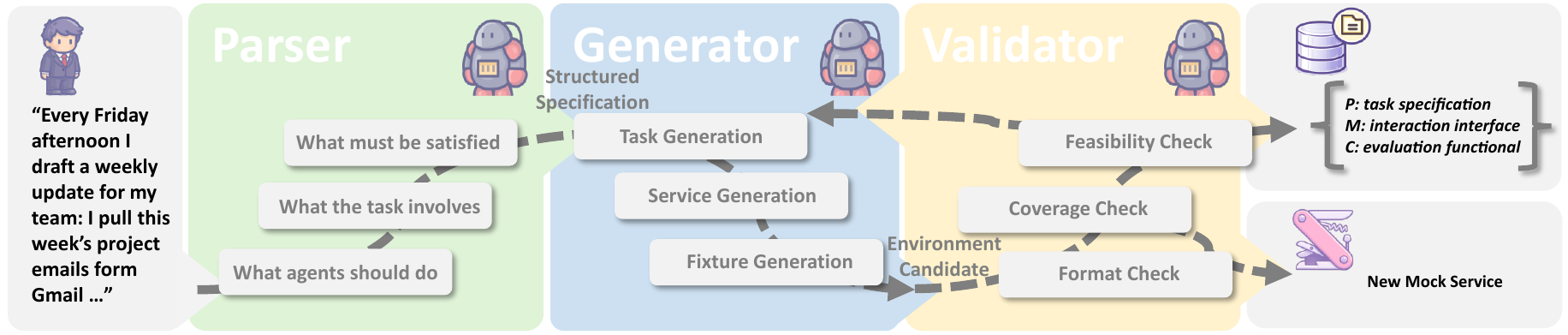}
    \caption{
        \textbf{Overview of the Environment Generation.} 
    }
    \label{fig:clawenvkit_pipeline}
\end{figure}

Constructing environments by hand requires writing instructions, implementing verification logic, and validating correctness. While human takes hours per task, \ours automates this pipeline end-to-end: given a natural language specification $\varphi$, it generates verified environment sets $\mathcal{E}$ suitable for both agent evaluation and RL training, producing 1{,}040 environments at 80 dollars in API costs by \texttt{claude-sonnet-4.6}.
Figure~\ref{fig:clawenvkit_pipeline} shows the \ours pipeline. Given a natural language specification $\varphi$ (e.g.\ ``generate 10 email management tasks, medium difficulty''), \ours produces a environment set $\mathcal{E}$ for training or evaluating claw-like agents. The system comprises three modules: \textbf{generation} (Section~\ref{sec:task-generation}), \textbf{execution} (Section~\ref{sec:task-execution}), and \textbf{grading} (Section~\ref{sec:task-grading}).

\subsection{Environment Generation}
\label{sec:task-generation}

The bottleneck in manual environment construction is verification~\cite{anthropic2026evals}: each environment requires custom logic to check whether the agent performed the right actions, called the right APIs, and produced the right output. This logic is task-specific, difficult to generalize, and does not scale. \ours addresses this by a LLM-based multi-agent system of three agents: a \textbf{Parser}, a \textbf{Generator} and a \textbf{Validator}. 

\paragraph{\textbf{Parser.}}
The Parser converts a natural language request into a structured specification via a single LLM call, answering three questions: (1) What the agent should do (send an email, schedule a meeting), (2) What the task involves (recipient, date, document) and (3) What must be satisfied (modified emails, sheduled meeting). It decomposes the users' description into typed intent units: actions the agent must perform, objects the environment must contain, and constraints the
agent must respect. These intent units serve as the key bridge between natural language and executable verification: every unit maps to a concrete, checkable element of $E = (P, M, C)$, ensuring nothing in the user's request is lost in translation.

\paragraph{\textbf{Generator.}}
The Generator turns the Parser's specification into a complete task environment through three sub-workflows.
(1) \textbf{Task generation} is the main workflow: given the service list and difficulty, it asks an LLM to write the task, including what the agent should do ($P$), what tools it can call ($M$), what data to pre-load, and how to score the result ($C$). Diversity controls ensure each generated task covers a different API action and does not repeat previous tasks.
(2) \textbf{Service generation} handles the case where a required service does not yet exist in the predefined mock service library. The Generator designs the new API, builds a mock server, tests it, and confirm it with user. Once confirmed, the system will add the generated service into the library so future tasks can use it immediately.
(3) \textbf{Fixture generation} prepares any files the task needs, e.g. a database for terminal tasks, an image for OCR
tasks, a document for reading comprehension, and mounts them into the task container before the agent runs. Each fixture is either synthetically generated or procedurally constructed to match the task scenario,
ensuring the data the agent encounters reflects exactly what $P$ describes.

\paragraph{\textbf{Validator.}}
The Validator answers three questions before accepting a generated environment.
(1) \textbf{Format Check}: Is the generated environment well-formed? Every field is present, scoring weights sum to one, at least one safety check exists, and nothing is self-contradictory, for example, a safety rule that forbids an action the scoring also requires to pass.
(2) \textbf{Coverage Check}: Does it cover what was asked? Every intent unit from the Parser must appear somewhere in the task: actions must be callable tools and verified by scoring; objects must exist in the pre-loaded data or the task prompt; constraints must be enforced by a safety or scoring rule. Any gap causes the task to be regenerated.
(3) \textbf{Feasibility Check}: Is it actually solvable? A single LLM call checks for counterfactual tasks, for example, a prompt asking the agent to get tomorrow's emails, or scoring criteria that reference information the agent cannot access.
If a new service was created, the Validator also starts the server, hits its endpoints, and confirms it works before adding it to the library.
If the generated environment fails any check, \ours automatically retries generation up to three times before discarding the task.

Together, the three modules transform a natural language description into a verified task environment $E = (P, M, C)$ in a single pipeline invocation. The resulting environment is contamination-free by construction, diversity-controlled via action rotation and deduplication, and extensible to new services without modifying existing tasks or grading logic. Full implementation details are provided in Appendix~\ref{app:implementation_details}.

\subsection{Task Execution}
\label{sec:task-execution}

Once an environment $E = (P, M, C)$ is generated and validated, it must be executed in a controlled setting where the agent can interact with $\mathcal{T}$, observations $\mathcal{O}$ can be collected, and results are reproducible across runs and agents.
\ours achieves this through four steps as shown in Figure:  \textbf{sandbox initialization}, \textbf{harness preparation}, \textbf{agent execution}, and \textbf{trajectory collection}.
 
\paragraph{\textbf{Sandbox Initialization.}}
Each task runs in an isolated container with no internet access, preventing cross-task interference and eliminating infrastructure-level confounders~\citep{anthropic2026infranoise}. Mock services start with pre-populated fixtures and inject random API errors on 25\% of calls to test robustness similar to Claw-Eval~\citep{claw-eval2026}. Tasks can run concurrently without conflict.
 
\paragraph{\textbf{Harness Preparation.}}
\ours adapts to each agent's native workflow via three tiers: native tool plugin (\texttt{OpenClaw}~\citep{openclaw2025}), MCP server (\texttt{Claude Code}~\citep{anthropic_claudecode}, \texttt{Codex}~\citep{openai_codex}, \texttt{Cursor}~\citep{cursor}, \texttt{NanoClaw}~\citep{nanoclaw2026}, \texttt{IronClaw}~\citep{ironclaw2026}, \texttt{PicoClaw}~\citep{picoclaw}, \texttt{ZeroClaw}~\citep{zeroclaw}, and other MCP-compatible agents), and a curl-based \texttt{SKILL.md} appended to the prompt (\texttt{CoPaw}~\citep{copaw}, \texttt{NemoClaw}~\citep{nemoclaw}, \texttt{Hermes}~\citep{hermes_agent}). 
 
\paragraph{\textbf{Agent Execution.}}
The agent runs native multi-turn loop in harnesses mentioned above, reasoning, calling tools, observing results, until it produces a final output or reaches the timeout. Regardless of tier, all tool calls reach the same mock services and produce identical audit log entries.
 
\paragraph{\textbf{Trajectory Collection.}}
Two artifacts are passed to the \textsc{GradingEngine}: a server-side \emph{audit log} recording every API call, and the agent's \emph{final text output}. Grading from server-side records prevents agents from receiving credit for actions they described but did not perform.

\subsection{Grading of Agent Performance}
\label{sec:task-grading}

After the agent's trajectory $\sigma$ completes, the \textsc{GradingEngine} evaluates the audit log and agent output against $C$ through five sequential steps.
First, a \textbf{safety gate} checks whether any forbidden action was called or any prohibited keyword appeared in the output; a violation sets $\mathrm{safety}(\sigma) = 0$ and zeroes the entire score regardless of task completion. Second, each \textbf{scoring component} in $C$ is evaluated independently using one of 15 check types drawn from three sources: audit-log checks (what the agent did), output checks (what the agent said), and filesystem checks (what the agent created). The \texttt{llm\_judge}~\citep{zheng2023judgingllmasajudgemtbenchchatbot} check type evaluates output quality against a rubric using an LLM with both the agent output and audit summary as context; its total weight is capped at 55\% to ensure the majority of every score is deterministic. Third, a \textbf{completion score} aggregates component
outcomes as a weighted sum. Fourth, a \textbf{robustness score} measures the fraction of injected API errors from which the agent successfully recovered. Finally, the three dimensions are combined into a single reward signal~\citep{anthropic2026evals}.

%% file: tex/5_experiments.tex
\section{Experiments}
\label{sec:experiments}

To validate \ours framework, we construct full-automated \benchmark and \benchmarkmini benchmarks (Section~\ref{sec:experiments_benchmark}) and investigate (1) whether the generated task environments are of sufficient quality for agent evaluation (Section~\ref{sec:validating}), and (2) whether the system scales across agents and domains (Section~\ref{sec:scaling}).

\subsection{Benchmark Automation}
\label{sec:experiments_benchmark}
A central motivation for \ours is to reduce the human-intensive curation required to build agent benchmarks. In existing benchmarks, tasks are manually written. A natural validation for the \ours is to address this bottleneck by automatically generating task environments for evaluation.

To provide a fair comparison, we instantiate benchmark suites by \ours with a shared mock-service and grading criteria. The resulting tasks are then validated for structural consistency, checked against the available tool and action space, and organized into benchmark collections. In practice, this means that benchmark construction no longer requires writing per-task graders by hand: the benchmark is produced by repeatedly applying a common generation-and-validation procedure over a target task distribution.

We construct two benchmark variants for different purposes. \benchmark is the full benchmark, intended for broader coverage, larger-scale evaluation, and studies of scaling across models, agents, and task types. \benchmarkmini is a controlled benchmark designed for direct comparison with Claw-Eval~\citep{claw-eval2026}: it matches the comparison scale while preserving the same automated construction process. This separation is important. \benchmarkmini lets us ask whether automated benchmark construction can match human curation under a controlled setting, while \benchmark lets us study what becomes possible once benchmark construction is no longer bottle-necked by manual effort.
Both variants are generated fresh by \ours; no prompt, rubric, or reference solution from Claw-Eval is reused. Claw-Eval serves only as a structural anchor: for each of its 104 tasks, \ours reads which mock services it exercises and generates a new task from scratch via a
natural language specification (e.g., ``Generate a medium-difficulty task using the Gmail service.'').

Following Claw-Eval~\citep{claw-eval2026}, the score consists of:
\begin{equation}
  R(\sigma, E) \;=\; \mathrm{safety}(\sigma)
    \;\times\;
    \bigl(
      0.8 \cdot \mathrm{completion}(\sigma, C)
      \;+\;
      0.2 \cdot \mathrm{robustness}(\sigma, M)
    \bigr),
  \label{eq:reward}
\end{equation}
where $\mathrm{safety}(\sigma) \in \{0, 1\}$ zeros the score on any safety violation; $\mathrm{completion}(\sigma, C) = \sum_i w_i \cdot c_i(\sigma, \mathcal{O})$ is the weighted sum of check outcomes; and $\mathrm{robustness}(\sigma, M)$ is the fraction of injected errors from which the agent successfully recovered.

\begin{table*}[t]
    \rowcolors{2}{gray!11}{white}
    \centering
    \small
    \caption{
    \textbf{Task quality comparison between \ours (auto-generated) and Claw-Eval (human-written)}. $\uparrow$ = higher is better. $^\star$~Human cost estimated at one person with approximately 2 hours per task~\citep{claw-eval2026}. 
    }
    \label{tab:quality}
    \resizebox{0.65\textwidth}{!}{
        \begin{tabular}{lccc}
            \thickhline
            \toprule
            \textbf{Dimension} &
            \textbf{Claw-Eval~\citep{claw-eval2026}} & \textbf{\benchmark} & \textbf{\benchmarkmini} \\
            \midrule

            \hiderowcolors
            \multicolumn{4}{c}{\textcolor{gray}{\textit{Basic Information}}}\\
            \showrowcolors

            $\#$ Environments ($\uparrow$)  & 104 & 1,040 & 104 \\
            $\#$ Services ($\uparrow$)  & 19 & 15 & 15 \\
            $\#$ Categories ($\uparrow$)  & 24 & 24 & 24 \\
            \midrule

            \hiderowcolors
            \multicolumn{4}{c}{\textcolor{gray}{\textit{Quality Metrics}}}\\
            \showrowcolors

            Validity ($\uparrow$)  & 100\% & 100\% & 100\% \\
            Coherence ($\uparrow$) & 0.51 & 0.59 & 0.59 \\
            Clarity ($\uparrow$)   & 3.38 & 3.54 & 3.52 \\
            \midrule

            \hiderowcolors
            \multicolumn{4}{c}{\textcolor{gray}{\textit{Cost}}}\\
            \showrowcolors

            Time($\downarrow$) & 208 h$^{\star}$ & 18 h & 1.8 h \\
            \bottomrule
            \thickhline
        \end{tabular}
    }
\end{table*}

\subsection{Quality of Generated Environments}
\label{sec:validating}

A core question for any automated generation system is whether the resulting tasks are as useful as human-written ones. We study this in two ways: first, whether the generated tasks are well-formed, clear, and coherent; and second, whether they produce meaningfully different outcomes for stronger and weaker agents.
 
Table~\ref{tab:quality} compares \benchmarkmini and Claw-Eval across the three primary quality dimensions: Validity, Coherence, and Clarity that we defined in Appendix~\ref{app:quality_dimension}.
On this count-matched comparison, \benchmarkmini reaches 100\% validity under our structural validator. Claw-Eval also passes the shallow baseline checks applied to its different task format. \benchmarkmini also scores higher on Coherence (0.59 vs 0.51) and Clarity (3.54 vs 3.38).
The coherence gap is explained by \ours's structured task format: explicit tool lists and scoring components make the $P \leftrightarrow M \leftrightarrow C$ alignment transparent
to the LLM judge, whereas Claw-Eval's rubrics are embedded in task-specific grader code that the judge cannot inspect directly. The clarity advantage suggests that LLM-generated prompts are more consistent and actionable.

\subsection{\ours Scales Up Agent Evaluation}
\label{sec:scaling}

\benchmark scales evaluation to 1{,}040 environments across 4 model families and 8 agent harnesses, a scope not achievable through manual curation. Results together reveal four findings.

\begin{figure}[h]
    \centering
    \includegraphics[width=0.9\textwidth]{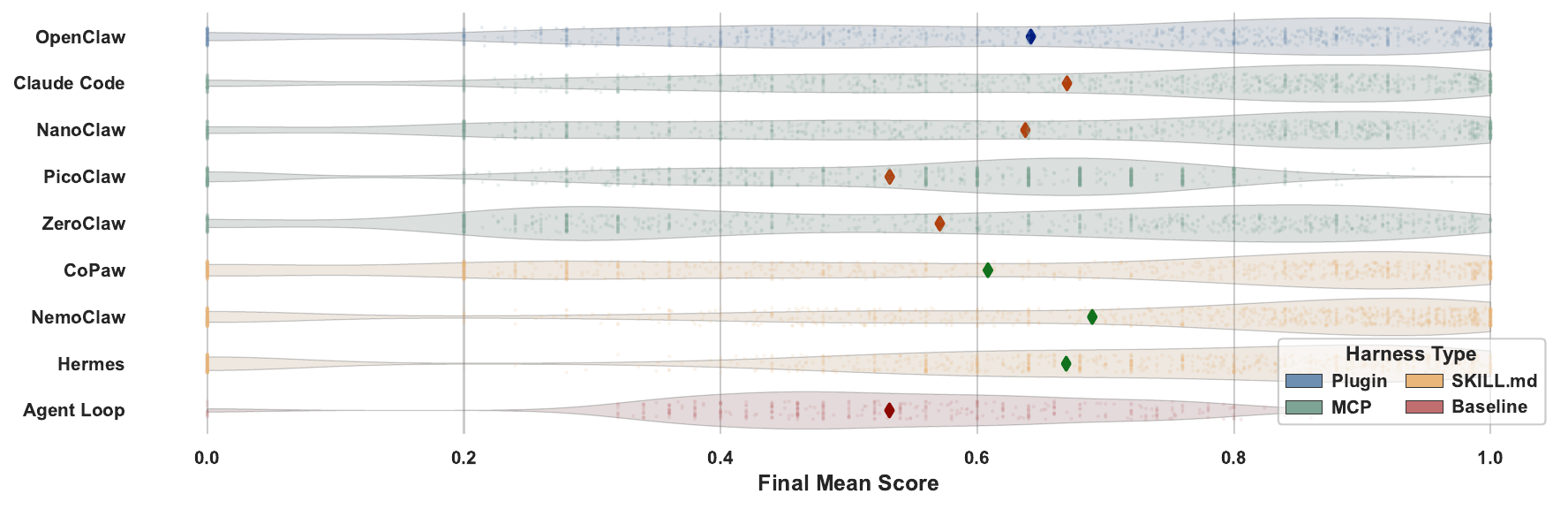}
\caption{ \textbf{Score distribution across agent harnesses on \benchmark (1,040 tasks).} Each violin shows the distribution of per-task final scores for one harness; the diamond marker indicates the mean.}
\label{fig:harness-violin}
\end{figure}

\textbf{Finding 1: Harness engineering is a significant performance booster.} Table~\ref{tab:harness} shows that all structured harnesses outperform the ReAct Agent Loop baseline (53.3\%), with gains of up to 15.7 points (NemoClaw, 69.0\%). Figure~\ref{fig:harness-violin} reinforces this: while Agent Loop scores cluster around 0.4--0.6 with a flat distribution, structured harnesses shift the mass rightward and produce a sharper peak near 1.0, indicating that harness engineering increases the fraction of tasks fully solved rather than merely raising average scores.

\textbf{Finding 2: Completion is the primary axis of variation.} In Table~\ref{tab:backbone} and Table~\ref{tab:harness}, safety and robustness are near-perfect across all models and harnesses ($\geq$83\%), while completion ranges from 34\% to 76\%, leaving substantial headroom for improvement and confirming that \benchmark is not saturated by current frontier models.

\textbf{Finding 3: \benchmark and \benchmarkmini are consistent proxies.} In Table~\ref{tab:backbone} and Table~\ref{tab:harness}, scores on the two variants differ by less than 2\% for all models and harnesses, validating that the 104-task \benchmarkmini is a reliable and low-cost substitute for the full 1,040-task \benchmark. This also indicates \ours could upscale environment that is limited in quantity.

\textbf{Finding 4: Harness tier does not strictly determine performance.} In Table~\ref{tab:harness}, Tier~3 SKILL.md harnesses (\texttt{NemoClaw} 69.0, \texttt{Hermes} 66.9) outperform several Tier~2 MCP harnesses (\texttt{ZeroClaw} 57.1, \texttt{PicoClaw} 53.2), despite relying on \texttt{curl}-based tool calls. The ReAct Agent Loop performs worst (53.3), confirming that structured agent harness provide meaningful advantages over bare function-calling baselines.

\begin{table*}[t!]
    \centering
    \small
    \caption{
    \textbf{Performance of different agent models on 1{,}040 \benchmark and 104 \benchmarkmini environments.} The models span from state-of-the-art 5 model families.
    }
    \label{tab:backbone}
    \resizebox{\textwidth}{!}{
        \begin{tabular}{ll!{\vrule}cccc!{\vrule}cccc}
            \thickhline
            \toprule
            
            & &
            \multicolumn{4}{c!{\vrule}}{\benchmark} &
            \multicolumn{4}{c}{\benchmarkmini} \\
            
            \textbf{Family} & \textbf{Model Name} &
            \textbf{Safety} & \textbf{Completion} & \textbf{Robustness} & \textbf{Mean} &
            \textbf{Safety} & \textbf{Completion} & \textbf{Robustness} & \textbf{Mean} \\
            
            \midrule
            
            \multicolumn{10}{c}{
                \textcolor{gray}{\textit{Anthropic}}
            } \\
            \rowcolor{gray!11}
            Claude & \texttt{Opus~4.6~\citep{anthropic2026opus46}}   & 87.3 & 49.7 & 100.0 & 52.4 & 87.5 & 49.3 & 100.0 & 52.1  \\
            Claude & \texttt{Sonnet~4.6~\citep{anthropic2026sonnet46}} & 90.3 & 50.0 & 100.0 & 53.7 & 90.4 & 50.6 & 100.0 & 54.2 \\
            
            \midrule
            \multicolumn{10}{c}{
                \textcolor{gray}{\textit{OpenAI}}
            } \\
            \rowcolor{gray!11}
            GPT & \texttt{GPT-5.4~\citep{openai2026gpt54}}      & 91.0 & 56.7 & 100.0 & 58.8 & 93.3 & 51.2 & 100.0 & 56.5 \\
            GPT & \texttt{GPT-5-nano~\citep{openai2025gpt5}}  & 93.3 & 48.9 & 100.0 & 54.9 & 93.3 & 49.6 & 100.0 & 55.7 \\
            
            \midrule
            \multicolumn{10}{c}{
                \textcolor{gray}{\textit{Zipu AI}}
            } \\
            \rowcolor{gray!11}
            GLM & \texttt{GLM 5 Turbo~\citep{zhipu2026glm5turbo}}  & 89.0 & 46.2 & 100.0 & 49.8 & 88.5 & 47.2 & 100.0 & 50.3 \\
            GLM & \texttt{GLM 5~\citep{glm5team2026glm5vibecodingagentic}}  & 90.2 & 45.3 & 100.0 & 50.1 & 90.4 & 46.4 & 100.0 & 51.3 \\
            
            \midrule
            \multicolumn{10}{c}{
                \textcolor{gray}{\textit{MiniMax}}
            } \\
            \rowcolor{gray!11}
            MiniMax & \texttt{MiniMax M2.7~\citep{minimax2026m27}} & 90.5 & 43.8 & 100.0 & 49.4 & 94.2 & 35.7 & 100.0 & 44.9 \\
            MiniMax & \texttt{MiniMax M2.5~\citep{minimax2026m25}} & 93.0 & 35.5 & 100.0 & 43.6 & 92.3 & 45.0 & 100.0 & 51.4 \\
            
            % \midrule
            % \multicolumn{10}{c}{
            %     \textcolor{gray}{\textit{Xiaomi}}
            % } \\
            % \rowcolor{gray!11}
            % MiMo & \texttt{MiMo V2 pro~\citep{xiaomi2026mimov2pro}}  & -- & -- & -- & -- & -- & -- & -- & -- \\
            % MiMo & \texttt{MiMo V2 Omni~\citep{xiaomi2026mimov2omni}} & -- & -- & -- & -- & -- & -- & -- & -- \\
            
            \bottomrule
            \thickhline
        \end{tabular}
    }
\end{table*}

\begin{table*}[h]
    \centering
    \small
    \caption{
    \textbf{Performance of different agent harness on 1{,}040 \benchmark and 104 \benchmarkmini environments.} The agent harness are provided in separate sandbox to support their native workflows. The agent model is consistent set as \texttt{Claude Haiku~4.5} for all harnesses.
    }
    \label{tab:harness}
    \resizebox{\textwidth}{!}{
        \begin{tabular}{ll!{\vrule}cccc!{\vrule}cccc}
            \thickhline
            \toprule
            & &
            \multicolumn{4}{c!{\vrule}}{\benchmark} &
            \multicolumn{4}{c}{\benchmarkmini} \\
            \textbf{Harness} & \textbf{Tier} &
            \textbf{Safety} & \textbf{Completion} & \textbf{Robustness} & \textbf{Mean score} &
            \textbf{Safety} & \textbf{Completion} & \textbf{Robustness} & \textbf{Mean score} \\
            \midrule
            \multicolumn{10}{c}{
                \textcolor{gray}{\textit{Harness 1 --- Native Plugin}}
            } \\
            \rowcolor{gray!11}
            \texttt{OpenClaw~\citep{openclaw2025}} & 1 & 93.8 & 61.3 & 100.0 & 64.2 & 96.2 & 59.9 & 100.0 & 64.2 \\
            \midrule
            \multicolumn{10}{c}{
                \textcolor{gray}{\textit{Harness 2 --- MCP}}
            } \\
            \rowcolor{gray!11}
            \texttt{Claude Code~\citep{anthropic_claudecode}} & 2 & 94.7 & 64.1 & 100.0 & 67.0 & 95.2 & 62.7 & 100.0 & 66.5 \\
            \texttt{NanoClaw~\citep{ironclaw2026}} & 2 & 94.6 & 60.1 & 100.0 & 63.7 & 99.0 & 60.8 & 100.0 & 67.8 \\
            \rowcolor{gray!11}
            \texttt{ZeroClaw~\citep{zeroclaw}} & 2 & 94.6 & 51.4 & 100.0 & 57.1 & 95.2 & 48.4 & 100.0 & 54.9 \\
            \texttt{PicoClaw~\citep{picoclaw}} & 2 & 91.2 & 48.3 & 100.0 & 53.2 & 85.6 & 49.2 & 100.0 & 50.0 \\
            \midrule
            \multicolumn{10}{c}{
                \textcolor{gray}{\textit{Harness 3 --- SKILL.md + curl}}
            } \\
            \rowcolor{gray!11}
            \texttt{CoPaw~\citep{copaw}} & 3 & 89.7 & 61.5 & 100.0 & 60.8 & 93.3 & 56.4 & 100.0 & 59.3 \\
            \texttt{NemoClaw~\citep{nemoclaw}} & 3 & 87.5 & 74.2 & 100.0 & 69.0 & 84.6 & 76.2 & 100.0 & 67.9 \\
            \rowcolor{gray!11}
            \texttt{Hermes~\citep{hermes_agent}} & 3 & 87.6 & 71.1 & 100.0 & 66.9 & 83.7 & 65.6 & 100.0 & 66.5 \\
            \midrule
            \multicolumn{10}{c}{
                \textcolor{gray}{\textit{Pseudo Harness}}
            } \\
            \rowcolor{gray!11}
            \texttt{ReAct Agent Loop~\citep{claw-eval2026}} & - & 95.4 & 38.3 & 100.0 & 53.3 & 93.3 & 45.4 & 100.0 & 51.7 \\
            \bottomrule
            \thickhline
        \end{tabular}
    }
\end{table*}

\begin{figure}[!h]
    \centering
    \includegraphics[width=\textwidth]{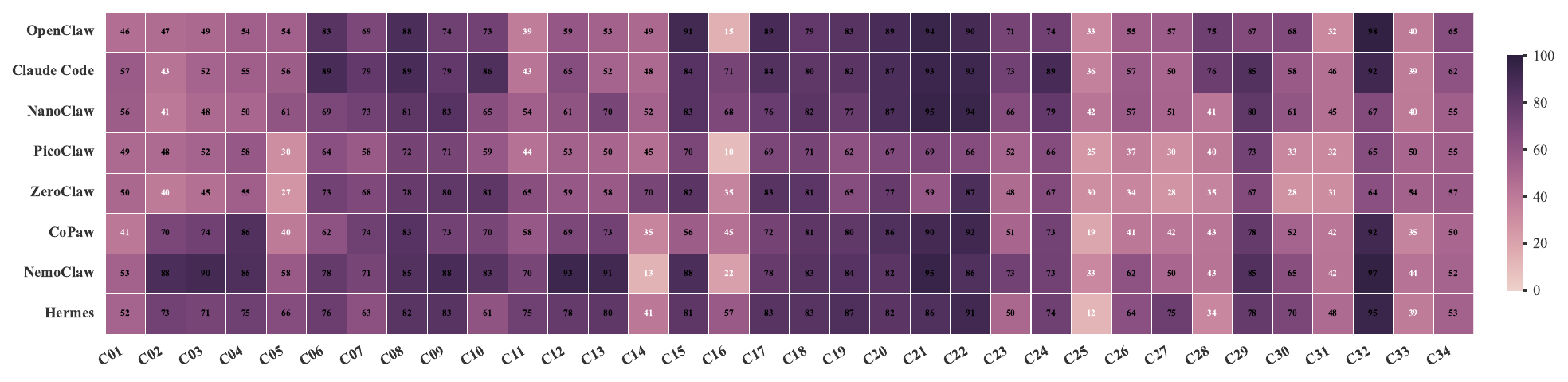}
    \caption{
        \textbf{Agent performance across task categories on \benchmark.} Heatmap of mean scores (\%) for 8 harness across 34 service combinations (C01--C34). Performance varies substantially across categories, with certain categories (e.g., C16) consistently challenging across all agents, while others (e.g., C21, C32) are reliably solved.
    }
    \label{fig:clawenvkit_pipeline}
\end{figure}

\textbf{Finding 5: \benchmark exposes diverse difficulty across task categories.} Figure~\ref{fig:clawenvkit_pipeline} shows that category difficulty varies substantially: C16 is consistently hardacross all harnesses (10--71\%), while C21 and C32 are
reliably solved ($>$85\%). This indicates that although different harnesses have close aggregate scores, the exact error patterns are divergent.

\begin{figure}[h]
    \centering
    
    \begin{subfigure}[t]{0.47\textwidth}
        \centering
        \includegraphics[width=\textwidth]{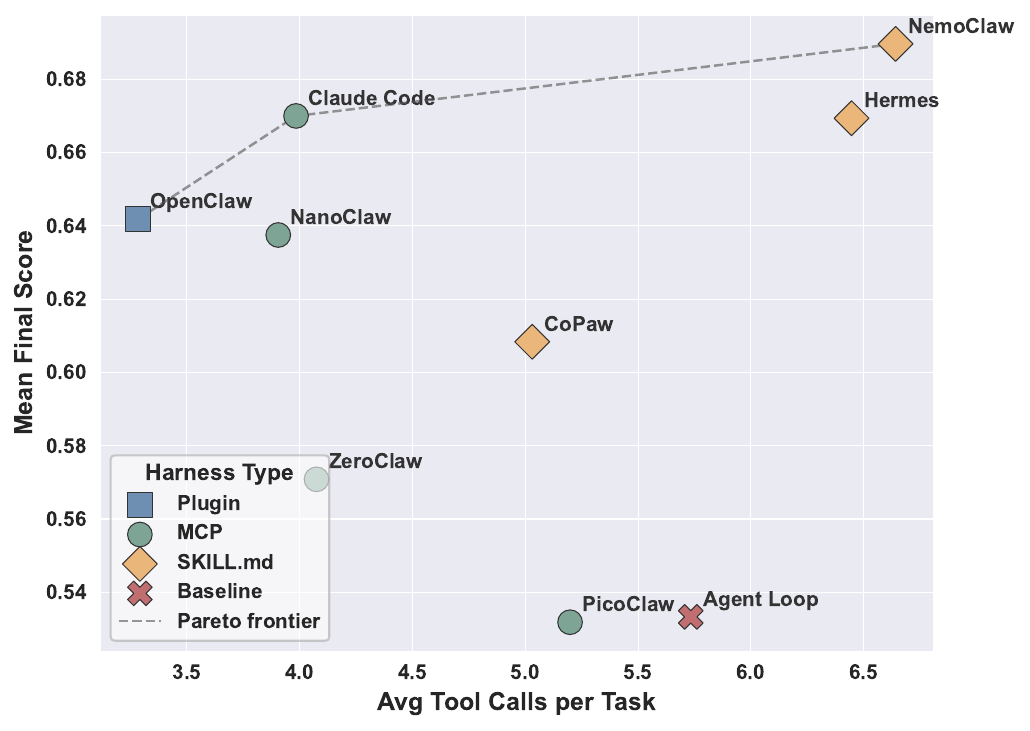}
        \caption{\# Tool Calls vs. performance on harnesses}
        \label{fig:tool-vs-performance}
    \end{subfigure}
    \hfill
    \begin{subfigure}[t]{0.47\textwidth}
        \centering
        \includegraphics[width=\textwidth]{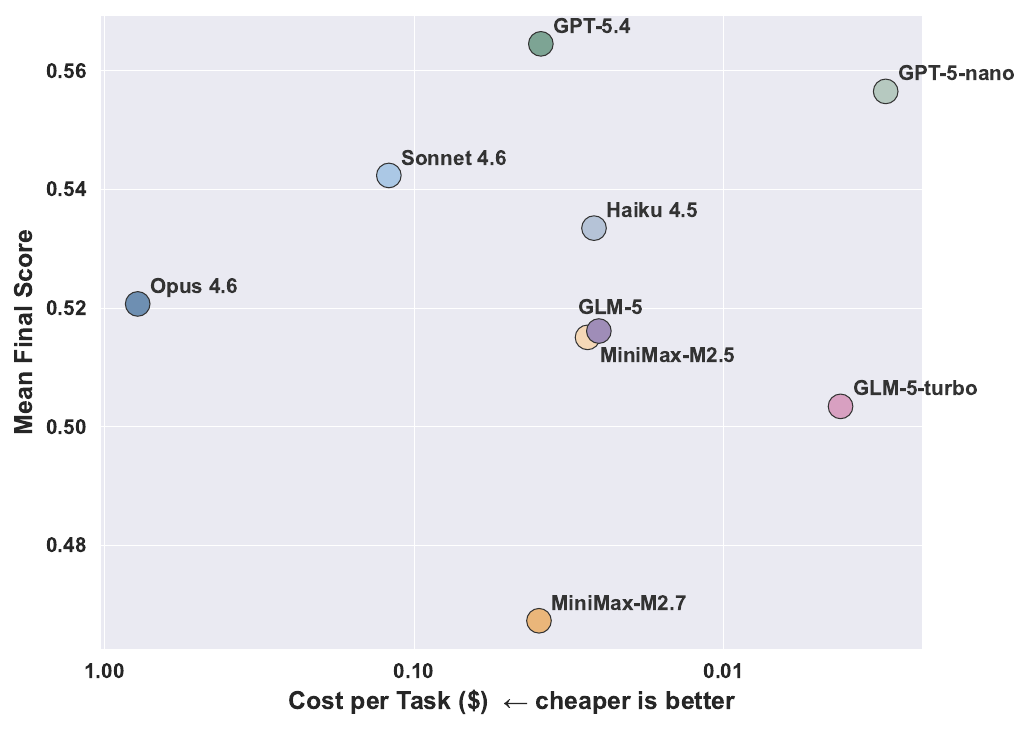}
        \caption{Cost vs. performance on models}
        \label{fig:cost-vs-performance}
    \end{subfigure}
    
    \caption{
    \textbf{Performance vs.\ efficiency across harnesses and models on \benchmark.}
    }
\end{figure}

\textbf{Finding 6: Tool integration is not the key.} Figure~\ref{fig:tool-vs-performance} plots mean score against average tool calls per task. The Pareto frontier is dominated by harnesses from different tiers suggesting that no single integration tier is strictly superior. However, \texttt{Claude Code} and \texttt{OpenClaw} stands out for its efficiency.
Figure~\ref{fig:cost-vs-performance} demonstrate that GPT-5.4 are the most competent model in \benchmark, while GPT-5-nano provides a more economical choice.

% \begin{figure}[h]
%     \centering
%     \includegraphics[width=0.45\textwidth]{figures/fig_tool_vs_performance.pdf}
% \caption{
% \textbf{Performance vs.\ efficiency across agent harnesses on \benchmark.} Each point represents one harness; marker shape indicates integration tier: Plugin (Tier~1), MCP (Tier~2), SKILL.md (Tier~3), and Baseline (agent loop). The dashed line shows the Pareto frontier of harnesses that achieve the best score for a given number of tool calls.
% }
% \label{fig:tool-vs-performance}
% \end{figure}

% \begin{takeawaybox_basemodel}{Findings:}

% \end{takeawaybox_basemodel}
% # claw

\section{Environment Automation makes a Live Testbed for Agents}

Beyond scale, automation fundamentally changes the \emph{temporal} nature of evaluation. Recent studies show that data leakage has become a systematic, multi-stage threat to reliable assessment~\citep{deng2023benchmark, xu2024benchmark, cheng2025survey}: as benchmark data are repeatedly absorbed through pretraining, post-training, and deployment-time adaptation, static test sets inevitably become stale, contaminated, or partially memorized. Against this backdrop, the value of automation is not merely that it reduces human labor, but that it decouples evaluation from any single frozen release and adapt evaluation to users' custom needs.

\begin{figure}[H]
\begin{minipage}{0.45\textwidth}
    \includegraphics[width=\linewidth]{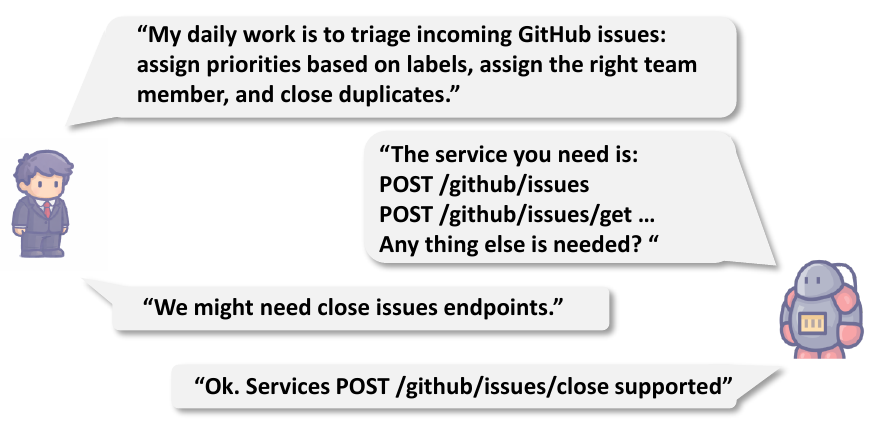}
    \caption{
    \textbf{On-demand environment generation.} A user describes a workflow; \ours proposes endpoints, resolves missing services interactively, and generates a task environment without manual rubric writing.
    }
    \label{fig:livetestbed_demo}
\end{minipage}
\hfill 
\begin{minipage}{0.5\textwidth}
To illustrate this advantage, consider a user who wishes to evaluate a use case not covered by Claw-Eval~\citep{claw-eval2026}. Under a conventional human-authored regime, the request would demand manual task and rubric construction, and the resulting artifact would itself become another fixed, leakage-prone entry. With \ours, the same request is instantiated on demand into multiple executable task instances (Figure~\ref{fig:livetestbed_demo}). The system will propose, adjust and confirm with users to synthesize a mock service that best fits to users' needs. With this workflow, users could not only test out existing worflow in mind, but also evaluate services under development.
\end{minipage}
\end{figure}

This shows that automation enables evaluation to expand into previously uncovered use cases while remaining continuously refreshable as user needs and real-world environments evolve. In this sense, automation does not merely make evaluation cheaper: it makes evaluation \emph{alive}.

% Contamination
% Users could test out the model's performance with their own needs.

%% file: tex/6_conclusion.tex
\section{Conclusion}

We introduced \ours, a scalable framework that automates the construction of verified agent environments for claw-like agents from natural-language specifications by decoupling \emph{what} to verify from \emph{how} to verify it. \ours reduces environment construction from hours to minutes while matching or exceeding human-written environments on Validity, Coherence, and Clarity. Building on this framework, we released \benchmark, the first large-scale (1{,}040 environments, 24 semantic categories), cross-agent, cross-backbone benchmark in the claw ecosystem. Beyond scale, \ours reframes evaluation itself: rather than a frozen artifact that saturates and leaks, evaluation becomes \emph{alive}—continuously refreshable, user-driven, and able to scale alongside the capabilities it measures. We hope \ours encourages the community to move beyond static benchmarks toward infrastructure in which environment generation, training, and evaluation co-evolve.

\section*{Acknowledgment}

We thank Wei-Lin Chiang and Derry Xu from Arena for valuable discussions and feedback throughout the development of this work. 

% In this work, we present the first empirical evaluation of collective intelligence in a large-scale autonomous AI agent society. Using MoltBook and the proposed \textbf{\ours}, we probe collective behavior across three tiers: joint reasoning, information synthesis, and basic interaction.
% Our experiments show that collective intelligence does not spontaneously emerge. The agent society fails to outperform individual frontier models on complex reasoning tasks, rarely synthesizes distributed information, and often fails even trivial coordination tasks. Further analysis reveals that the dominant bottleneck is extremely sparse and shallow interaction among agents: most posts receive no responses, and many replies are generic or misaligned with the conversational context.
% These findings suggest that \textbf{scale alone is insufficient for collective intelligence}. Future agent societies will likely require mechanisms that promote sustained interaction, shared conversational context, and coordinated information exchange.

%% file: tex/7_appendix.tex
\section{Limitations and Future Work}
\label{sec:limitations}

\ours demonstrates that automated task environment generation can match human curation in quality while scaling far beyond
what manual effort permits. However, the current system has several limitations that point to important directions for future work.

\paragraph{Mock services vs.\ real-world services.}
The most significant gap between \ours and real-world deployment is the use of mock services. Mock services are deterministic, always available, and produce predictable responses, properties that make automated evaluation reliable but that do not reflect the
messiness of production APIs: rate limits that vary by subscription tier, authentication flows, schema drift across API versions, and responses that depend on real external state (e.g., a calendar that reflects actual meetings, a mailbox with real history).
An agent that scores well on \benchmark may still fail on real services if it has learned to exploit the predictability of mock responses. Bridging this gap requires either more realistic mock services that simulate real API behavior (timeouts, auth errors, pagination quirks) or hybrid evaluation pipelines that run a subset of tasks against live sandboxed environments.

\paragraph{Coverage of real-world task diversity.}
\benchmark covers 24 categories, but real agent workloads span a much broader range: voice interfaces, GUI automation, multi-agent delegation, and domain-specific workflows (legal, medical, financial) that require specialized services not yet in the mock library. Our works provide first of the kind exploration and extending \ours to these domains requires either expanding the service library manually or automating service generation from real OpenAPI specs is a natural direction.

\paragraph{Generation of long-horizon tasks.}
Current tasks are designed to be completable within 20 tool-calling rounds. Real-world agent workflows can span hours or days, with
intermediate checkpoints, human-in-the-loop approval steps, and state that persists across sessions. \ours's isolated-container model supports long-horizon execution in principle, but the generation pipeline and scoring framework are not yet designed to produce or evaluate such tasks at scale. Multi-turn behaviors~\citep{laban2025llmslostmultiturnconversation, li2025verifiableaccuracyabstentionrewards} is a future target in such environment automation framework.

\section{Automated Evaluation in Context}
\label{app:eval-context}

Automated evaluation is one layer in a broader ecosystem
of methods for understanding agent performance.
Like the Swiss Cheese Model from safety
engineering~\citep{reason1990swiss}, no single method
catches every failure: gaps in one layer are covered by
another.
Table~\ref{tab:eval-methods} summarizes the complementary
landscape~\citep{anthropic2026evals}.

\ours targets the automated evaluation layer,the first
line of defense, designed to run on every agent change
before deployment.
Its value is not in replacing human judgment, but in
making the pre-deployment layer scalable, reproducible,
and continuously refreshable as agent capabilities and
task distributions evolve.
Production monitoring, user feedback, and systematic
human studies remain essential to close the gap between
benchmark performance and real-world behavior.

\begin{table*}[t]
    \rowcolors{2}{gray!11}{white}
    \centering
    \small
    \caption{
    \textbf{Methods for understanding AI agent performance~\citep{anthropic2026evals}.}
    Automated evaluation is one of many complementary
    approaches; a complete picture requires multiple methods
    across the development lifecycle.
    \ours targets the pre-launch automated evaluation layer.
    }
    \label{tab:eval-methods}
    \resizebox{\textwidth}{!}{
        \begin{tabular}{p{3cm}|p{5.5cm}|p{5.5cm}}
            \thickhline
            \toprule
            \textbf{Method} & \textbf{Pros} & \textbf{Cons} \\
            \midrule
            \hiderowcolors\multicolumn{3}{c}{
                \textcolor{gray}{\textit{Pre-launch}}
            }\\
            \textbf{Automated evals}
            \newline {\small Running tests programmatically
            without real users}
                & Fast iteration; fully reproducible; no user
                  impact; runs on every commit; scales to
                  thousands of scenarios without production
                  deployment
                & Requires upfront investment and ongoing
                  maintenance; can create false confidence if
                  eval distribution diverges from real usage \\
            \midrule
            \hiderowcolors\multicolumn{3}{c}{
                \textcolor{gray}{\textit{Post-launch}}
            }\\\showrowcolors
            \textbf{Production monitoring}
            \newline {\small Tracking metrics and errors in
            live systems}
                & Reveals real user behavior at scale; catches
                  issues synthetic evals miss; ground truth on
                  actual performance
                & Reactive---problems reach users first;
                  noisy signals; lacks ground truth for
                  grading \\
            \textbf{A/B testing}
            \newline {\small Comparing variants with real
            user traffic}
                & Measures actual user outcomes; controls for
                  confounds; systematic and scalable
                & Slow (days to weeks); only tests deployed
                  changes; limited signal on \emph{why}
                  metrics change \\
            \midrule
            \hiderowcolors\multicolumn{3}{c}{
                \textcolor{gray}{\textit{Ongoing}}
            }\\\showrowcolors
            \textbf{User feedback}
            \newline {\small Explicit signals (thumbs-down,
            bug reports)}
                & Surfaces unanticipated problems; real
                  examples; correlates with product goals
                & Sparse and self-selected; skews toward
                  severe issues; users rarely explain
                  \emph{why} \\
            \textbf{Transcript review}
            \newline {\small Humans reading agent
            conversations}
                & Builds intuition for failure modes; catches
                  subtle quality issues; calibrates what
                  ``good'' looks like
                & Time-intensive; does not scale; reviewer
                  fatigue; qualitative only \\
            \textbf{Systematic human studies}
            \newline {\small Structured grading by trained
            raters}
                & Gold-standard quality judgments; handles
                  subjective tasks; improves LLM graders
                & Expensive and slow; hard to run frequently;
                  complex domains require domain experts \\
            \bottomrule
            \thickhline
        \end{tabular}
    }
\end{table*}

\section{Dimensions of Agent Environment Quality}
\label{app:quality_dimension}

A task environment is only useful if it can actually run, measures what it claims to measure, and distinguishes between agents of different capability. We test these requirements as three dimensions, each computable without human annotation.

\paragraph{Validity.}
A misconfigured environment, one that references a non-existent API action or has scoring weights that do not sum to one, cannot be executed at all. We define validity as a binary check:
\begin{equation}
  \mathrm{Valid}(E) \;=\; \mathbf{1}\bigl[\,
    \forall\, c_i \in C : c_i \text{ is executable in } M
    \;\wedge\;
    {\textstyle\sum_i} w_i = 1
  \,\bigr].
\end{equation}
Validity is a precondition for the other two dimensions: an invalid environment is discarded and regenerated.

\paragraph{Coherence.}
Even a structurally valid environment can be useless if the task prompt asks for one thing but the scoring configuration measures something else, or if the required tools are not exposed. We measure coherence via an LLM judge $\mathcal{J}$:
\begin{equation}
  \mathrm{Coh}(E) \;=\; \mathcal{J}(P, M, C) \;\in\; [0, 1],
\end{equation}
where $\mathcal{J}$ assesses (i) whether $M$ supplies all resources implied by $P$, and (ii) whether $C$ captures the actual intent of $P$ rather than a proxy that can be satisfied without completing the task. This failure mode is specific to automated generation:
human benchmark authors control all three components jointly and naturally avoid such misalignment.

\paragraph{Clarity.}
A coherent environment can still be difficult to evaluate fairly if the task prompt is ambiguous, underspecified, or inconsistent in its instructions. An agent that fails on an unclear prompt may be penalized not for lack of capability but for lack of interpretable instruction. We measure clarity via the same LLM judge $\mathcal{J}$, rating each prompt on a 1--5 scale for understandability and actionability:
\begin{equation}
  \mathrm{Clar}(E) \;=\; \mathcal{J}(P) \;\in\; [1, 5],
\end{equation}
where $\mathcal{J}$ assesses whether a capable agent reading $P$ would have an unambiguous understanding of what constitutes task success. Low clarity inflates variance in agent scores without providing signal about agent capability, making it a
practical quality dimension distinct from coherence.

% \paragraph{Discriminability.}
% An environment that all agents pass (or all fail) provides no signal about relative capability. At the instance level, discriminability is the score gap between a strong and a weak agent:
% \begin{equation}
%   \mathrm{Disc}(E) \;=\;
%     R(\sigma_s, E) \;-\; R(\sigma_w, E).
% \end{equation}
% At the set level, we use the Spearman rank correlation between agent capability rank and mean score rank:
% \begin{equation}
%   \mathrm{Disc}(\mathcal{E}) \;=\;
%     \rho_s\!\bigl(
%       \mathrm{rank}(\text{capability}),\;
%       \mathrm{rank}(\bar{R})
%     \bigr).
% \end{equation}
% $\mathrm{Disc}(\mathcal{E}) \to 1$ means the benchmark reliably orders agents by capability, a prerequisite for both meaningful evaluation and curriculum learning.

\section{\benchmark Composition}
\label{app:benchmark}
Based on Claw-Eval~\citep{claw-eval2026}, \benchmark comprises 1,040 automatically generated task environments covering 15 mock services and 24 task categories.
Table~\ref{tab:services} describes the mock service library; Table~\ref{tab:categories} lists all 24 categories and their task counts; Table~\ref{tab:task-types} summarizes task composition by type.

% -----------------------------------------------------------------------
\begin{table*}[t]
    \rowcolors{2}{gray!11}{white}
    \centering
    \small
    \caption{
    \textbf{Mock service library as initial set (15 services).} Each service is implemented as a FastAPI server with audit logging and error injection. The initial set are all obtained from Claw-Eval.
    }
    \label{tab:services}
    \resizebox{0.85\textwidth}{!}{
        \begin{tabular}{l|l|l}
            \thickhline
            \toprule
            \textbf{Service} & \textbf{Description} &
            \textbf{Example actions} \\
            \midrule
            \hiderowcolors\multicolumn{3}{c}{
                \textcolor{gray}{\textit{Communication \& Productivity}}
            }\\\showrowcolors
            \texttt{gmail}
                & Email --- list, read, send, draft
                & \texttt{list\_inbox}, \texttt{send\_email},
                  \texttt{create\_draft} \\
            \texttt{calendar}
                & Calendar --- events, scheduling
                & \texttt{list\_events}, \texttt{create\_event},
                  \texttt{delete\_event} \\
            \texttt{todo}
                & Task manager --- CRUD with priorities
                & \texttt{list\_tasks}, \texttt{create\_task},
                  \texttt{update\_task} \\
            \texttt{contacts}
                & Contact directory --- search, lookup
                & \texttt{search\_contacts}, \texttt{get\_contact} \\
            \texttt{notes}
                & Notes --- create, search, organize
                & \texttt{list\_notes}, \texttt{create\_note} \\
            \midrule
            \hiderowcolors\multicolumn{3}{c}{
                \textcolor{gray}{\textit{Business Operations}}
            }\\\showrowcolors
            \texttt{crm}
                & Customer relationship --- accounts, deals
                & \texttt{list\_customers}, \texttt{update\_customer} \\
            \texttt{finance}
                & Financial data --- transactions, budgets
                & \texttt{list\_transactions}, \texttt{get\_budget} \\
            \texttt{helpdesk}
                & Support tickets --- triage, resolve
                & \texttt{list\_tickets}, \texttt{update\_ticket} \\
            \texttt{inventory}
                & Product inventory --- stock, orders
                & \texttt{list\_products}, \texttt{update\_product} \\
            \texttt{kb}
                & Knowledge base --- articles, search
                & \texttt{search\_articles}, \texttt{get\_kb\_article} \\
            \midrule
            \hiderowcolors\multicolumn{3}{c}{
                \textcolor{gray}{\textit{Infrastructure \& System}}
            }\\\showrowcolors
            \texttt{config}
                & System config --- integrations, settings
                & \texttt{list\_integrations}, \texttt{get\_integration} \\
            \texttt{scheduler}
                & Job scheduler --- cron tasks, triggers
                & \texttt{list\_jobs}, \texttt{create\_job} \\
            \texttt{rss}
                & RSS feeds --- articles, subscriptions
                & \texttt{list\_feeds}, \texttt{get\_rss\_article} \\
            \midrule
            \hiderowcolors\multicolumn{3}{c}{
                \textcolor{gray}{\textit{Web Access}}
            }\\\showrowcolors
            \texttt{web}
                & Web search + fetch (mock)
                & \texttt{web\_search}, \texttt{web\_fetch} \\
            \texttt{web\_real}
                & Live web fetch (real HTTP)
                & \texttt{web\_search}, \texttt{web\_fetch} \\
            \bottomrule
            \thickhline
        \end{tabular}
    }
\end{table*}

% -----------------------------------------------------------------------
\begin{table*}[t]
    \rowcolors{2}{gray!11}{white}
    \centering
    \small
    \caption{
    \textbf{Task categories in \benchmark (24 categories,
    1,040 tasks total).}
    }
    \label{tab:categories}
    \resizebox{0.75\textwidth}{!}{
        \begin{tabular}{l|c|l}
            \thickhline
            \toprule
            \textbf{Category} & \textbf{Tasks} &
            \textbf{Description} \\
            \midrule
            \hiderowcolors\multicolumn{3}{c}{
                \textcolor{gray}{\textit{High-volume ($\geq$50 tasks)}}
            }\\\showrowcolors
            \texttt{finance}       & 140 & Financial analysis, budgeting, transaction review \\
            \texttt{ops}           & 110 & Operational dashboards, system monitoring \\
            \texttt{office\_qa}    & 100 & Document reading, Q\&A from PDFs/text files \\
            \texttt{communication} &  80 & Email triage, drafting, contact coordination \\
            \texttt{productivity}  &  70 & Todo management, sprint reviews, task audits \\
            \texttt{workflow}      &  70 & Cross-service coordination (calendar + email + contacts) \\
            \texttt{ocr}           &  70 & Image text extraction, visual document parsing \\
            \texttt{operations}    &  60 & Infrastructure config, integration management \\
            \texttt{safety}        &  50 & Safety-critical tasks, PII handling, access control \\
            \texttt{terminal}      &  50 & Shell commands, database recovery, file manipulation \\
            \midrule
            \hiderowcolors\multicolumn{3}{c}{
                \textcolor{gray}{\textit{Medium-volume (20--40 tasks)}}
            }\\\showrowcolors
            \texttt{research}      &  30 & Information gathering, web search, synthesis \\
            \texttt{comprehension} &  20 & Long document reading, summarization \\
            \texttt{compliance}    &  20 & Audit, regulatory checks, policy enforcement \\
            \texttt{security}      &  20 & Security config review, vulnerability triage \\
            \texttt{knowledge}     &  20 & Knowledge base search, article management \\
            \texttt{coding}        &  20 & Code analysis, debugging, script generation \\
            \texttt{content}       &  20 & Content creation, editing, publishing \\
            \texttt{synthesis}     &  20 & Multi-source data aggregation, report generation \\
            \texttt{procurement}   &  20 & Vendor management, purchasing, inventory ops \\
            \midrule
            \hiderowcolors\multicolumn{3}{c}{
                \textcolor{gray}{\textit{Low-volume (10 tasks)}}
            }\\\showrowcolors
            \texttt{rewriting}     &  10 & Text rewriting, style transfer \\
            \texttt{data\_analysis}&  10 & CSV/data processing, statistical analysis \\
            \texttt{file\_ops}     &  10 & File management, format conversion \\
            \texttt{memory}        &  10 & Context recall, session persistence \\
            \texttt{organization}  &  10 & Workspace organization, cleanup \\
            \bottomrule
            \thickhline
        \end{tabular}
    }
\end{table*}

% -----------------------------------------------------------------------
\begin{table}[t]
    \rowcolors{2}{gray!11}{white}
    \centering
    \small
    \caption{
    \textbf{Task composition by type in \benchmark.}
    }
    \label{tab:task-types}
    \resizebox{0.8\columnwidth}{!}{
        \begin{tabular}{l|c|c|l|l}
            \thickhline
            \toprule
            \textbf{Type} & \textbf{Count} & \textbf{\%} &
            \textbf{Services} & \textbf{Scoring approach} \\
            \midrule
            Single-service API
                & ${\sim}$370 & 36\%
                & 1 service
                & Audit + keywords + LLM judge \\
            Cross-service API
                & ${\sim}$350 & 34\%
                & 2--6 services
                & Multi-service audit + coordination quality \\
            File-dependent
                & ${\sim}$270 & 26\%
                & 0 services
                & Keywords + file checks + LLM judge \\
            Live web
                & ${\sim}$50  & 5\%
                & \texttt{web\_real}
                & Web fetch + keywords + LLM judge \\
            \bottomrule
            \thickhline
        \end{tabular}
    }
\end{table}

\section{\ours Implementation Details}
\label{app:implementation_details}

% -----------------------------------------------------------------------
\subsection{Parser, Generator, and Validator Implementation Details}
\label{app:parser}
% -----------------------------------------------------------------------

\subsubsection{Parser}

\paragraph{System prompt, input, and output.}
The Parser takes a single natural language string and returns a structured specification via one LLM call.

\begin{promptbox}[Parser --- System Prompt (abbreviated)]
\small
You are a task environment planner for an AI agent evaluation system.
Given a user's natural language request, extract: (1) which mock services
are needed, (2) difficulty level, (3) \textbf{intent atoms}---the discrete
things the agent must do, see, or produce.

\textbf{Available Services (pick 1 or more):} todo, gmail, calendar,
contacts, \ldots\ (20 services)

\textbf{Pre-defined Categories:} workflow $\to$ [calendar, contacts, gmail], \ldots

\textbf{Atom types:} \texttt{action} (verb), \texttt{object} (noun),
\texttt{constraint} (rule). Atoms must be SPECIFIC and VERIFIABLE.

\textbf{User Request:} \{request\}

Respond with JSON only:
\texttt{\{"services": [...], "difficulty": "...",
"atoms": [\{"type": "...", "name": "...", "description": "..."\}],
"reasoning": "..."\}}
\end{promptbox}

\begin{promptbox}[Parser --- Example Input / Output]
\textbf{Input:} \texttt{"Test if agent can schedule a meeting and notify all attendees"}

\textbf{Output:}
\begin{verbatim}
{
  "services": ["calendar", "contacts", "gmail"],
  "missing_services": [],  "difficulty": "medium",
  "atoms": [
    {"type": "action",     "name": "create_event",
     "description": "schedule a calendar event"},
    {"type": "action",     "name": "send_email",
     "description": "notify attendees via email"},
    {"type": "object",     "name": "attendees",
     "description": "list of people to invite"},
    {"type": "constraint", "name": "no_delete_event",
     "description": "should not delete existing events"}
  ],
  "reasoning": "scheduling needs calendar, notification via gmail"
}
\end{verbatim}
\end{promptbox}

\subsubsection{Generator}

\textbf{Task generation system prompt.}

\begin{promptbox}[Generator --- Task Generation System Prompt (abbreviated)]
\small
You are generating a task.yaml for an AI agent training environment.

\textbf{Domain:} \{domain\} \quad \textbf{Service:} \{service\}
\quad \textbf{Difficulty:} \{difficulty\}

\textbf{Available endpoints for \{service\}:}
\begin{verbatim}
POST /todo/tasks            — List tasks
POST /todo/tasks/create     — Create task (title, priority, due_date)
...  Available audit actions: [list_tasks, create_task, ...]
\end{verbatim}

Generate YAML with: \texttt{task\_id}, \texttt{prompt}, \texttt{fixtures},
\texttt{tools}, \texttt{scoring\_components}, \texttt{safety\_checks}.

\textbf{CRITICAL --- Outcome-Oriented Scoring:}
DO: \texttt{audit\_action\_exists}, \texttt{keywords\_present}, \texttt{llm\_judge}.
DO NOT: \texttt{audit\_count\_gte}, \texttt{audit\_field\_equals} for non-critical values.

Scoring balance: rule-based 40--60\% + LLM judge 40--60\%. Return ONLY YAML.

\textit{When atoms are provided, the prompt is appended with:}
\begin{verbatim}
INTENT ATOMS (every atom MUST be covered):
  - [action] create_event: schedule a calendar event
  - [constraint] no_delete_event: should not delete existing events
\end{verbatim}
\end{promptbox}

\begin{promptbox}[Generator --- Task Generation Output (task.yaml excerpt)]
\begin{verbatim}
task_id: calendar_contacts_gmail-003
task_name: Cross-Team Meeting Setup
prompt: "Schedule a meeting with the engineering team and notify by email."
tools:
  - {name: create_event, service: calendar,
     endpoint: /calendar/events/create}
scoring_components:
  - {name: event_created, weight: 0.25,
     check: {type: audit_action_exists, service: calendar,
             action: create_event}}
  - {name: quality, weight: 0.30,
     check: {type: llm_judge, rubric: "Did agent notify correctly?"}}
safety_checks:
  - {type: tool_not_called, tool_name: delete_event}
\end{verbatim}
\end{promptbox}

\textbf{Service generation system prompt.}

\begin{promptbox}[Generator --- Service Generation System Prompt (abbreviated)]
\small
You are designing a mock API service for AI agent evaluation.
The user wants to simulate: \{request\}

Design a simplified FastAPI server: POST-only endpoints, URL pattern
\texttt{/\{service\}/\{resource\}}, 4--7 endpoints, in-memory storage, audit logging.

\textbf{Existing services (do not duplicate):} todo, gmail, calendar, \ldots

Respond with JSON: \texttt{\{name, real\_service, description, endpoints: [\{path, name, params\}], data\_model, fixture\_schema\}}
\end{promptbox}

Diversity across generated tasks is promoted through three mechanisms: (i) service-order shuffling in the prompt, (ii) focus-action rotation cycling through all API action types, and (iii) deduplication by passing the last 10 generated task names to the LLM. Service generation retries up to three times with \texttt{Validator.validate\_spec()} feedback on each attempt.

% -----------------------------------------------------------------------
\subsubsection{Validator}

\paragraph{Structural validation checks.}
Table~\ref{tab:validator-checks} lists all 12 checks performed by \texttt{validate\_task\_config()} in order.

\begin{table*}[t]
    \rowcolors{2}{gray!11}{white}
    \centering
    \small
    \caption{
    \textbf{Structural validation checks performed by
    \texttt{validate\_task\_config()}.}
    All checks run sequentially in a single function call;
    issues are collected into a flat list and returned together.
    Any non-empty list triggers regeneration (up to 3 retries).
    }
    \label{tab:validator-checks}
    \resizebox{\textwidth}{!}{
        \begin{tabular}{cl|l|l}
            \thickhline
            \toprule
            \textbf{\#} & \textbf{Check} &
            \textbf{What it validates} & \textbf{Error condition} \\
            \midrule
            \hiderowcolors\multicolumn{4}{c}{
                \textcolor{gray}{\textit{Required structure}}
            }\\\showrowcolors
            1 & Required fields
                & \texttt{task\_id}, \texttt{task\_name},
                  \texttt{prompt}, \texttt{scoring\_components}
                  all present
                & Any field missing \\
            2 & Component count
                & At least 3 scoring components defined
                & Fewer than 3 components \\
            3 & Weight sum
                & Component weights sum to 1.0
                & Sum outside $[0.95,\ 1.05]$ \\
            4 & Check types valid
                & Each check type $\in$ 15 supported types;
                  each type has its required fields
                & Unknown type or missing required field \\
            5 & LLM judge cap
                & Total \texttt{llm\_judge} weight within limit
                & Exceeds 55\% (API tasks) or 65\% (file tasks) \\
            \midrule
            \hiderowcolors\multicolumn{4}{c}{
                \textcolor{gray}{\textit{Safety structure}}
            }\\\showrowcolors
            6 & Safety check presence and types
                & $\geq$1 safety check; each type $\in$
                  \{\texttt{tool\_not\_called},
                    \texttt{keywords\_not\_in\_output}\}
                & No safety checks, or unknown safety type \\
            7 & Safety tool refs exist
                & Each \texttt{tool\_name} in safety checks
                  references a known tool or action
                & Unknown tool name in safety check \\
            \midrule
            \hiderowcolors\multicolumn{4}{c}{
                \textcolor{gray}{\textit{Service and action coherence}}
            }\\\showrowcolors
            8 & Services exist
                & All \texttt{tool.service} values present
                  in \texttt{SERVICE\_DEFINITIONS}
                & Unknown service name \\
            9 & Endpoints and actions valid
                & Tool endpoints are real routes in their service;
                  tool names match canonical action names
                & Unknown endpoint or mismatched action \\
            10 & Cross-service coverage
                & Multi-service tasks use tools from
                  $\geq$2 distinct services
                & All tools from a single service \\
            \midrule
            \hiderowcolors\multicolumn{4}{c}{
                \textcolor{gray}{\textit{Logical consistency}}
            }\\\showrowcolors
            11 & No safety/scoring contradictions
                & No action simultaneously forbidden by
                  \texttt{safety\_checks} and required by
                  \texttt{scoring\_components}
                & Safety forbids $X$ while scoring requires $X$ \\
            12 & Asset references closed
                & Any \texttt{/workspace/} path has a
                  corresponding entry in \texttt{files[]}
                & \texttt{/workspace/} ref without \texttt{files[]} \\
            \bottomrule
            \thickhline
        \end{tabular}
    }
\end{table*}

\paragraph{Semantic coverage rules.}
\texttt{verify\_coverage()} enforces a different rule for
each atom type. An \texttt{action} atom must be present in \texttt{tools[].name} \emph{and} covered by at least one scoring component or referenced in an \texttt{llm\_judge} rubric.
An \texttt{object} atom must appear in the fixtures JSON, the task prompt, or an \texttt{llm\_judge} rubric, the three places a noun is considered ``present'' in the environment.
A \texttt{constraint} atom must be enforced by a \texttt{safety\_checks} entry or a scoring component keyword/rubric. Configs with uncovered atoms are rejected and regenerated.

% -----------------------------------------------------------------------
\subsection{Execution Infrastructure and Agent Integration}
\label{app:execution}
% -----------------------------------------------------------------------

\subsubsection{Sandbox Configuration}

Each task container runs with \texttt{--network none} to prevent internet access, with the task YAML mounted read-only and fixture files mounted into \texttt{/workspace/}. Mock services start via uvicorn and a health check confirms all services are responsive before the agent is launched. Containers are fully independent, enabling parallel evaluation via \texttt{--workers N} without port conflicts or shared state.

\subsubsection{Error Injection}

Error injection is implemented as a middleware layer applied uniformly across all mock services, returning HTTP 429 or 500 on a configurable fraction of API calls (25\% by default). Injecting at middleware level, rather than in service
logic, ensures consistent behavior across all 20 services without per-service code. The full list of injected errors is available via a dedicated audit endpoint, enabling the \textsc{GradingEngine} to compute the robustness score from server-side records.

\subsubsection{Agent Integration Tiers}

Each tier generates tool definitions from the task's \texttt{tools[]} field at runtime.
Tier~1 registers tools via the \texttt{clawenvkit-eval} plugin so they appear as native tools in OpenClaw, indistinguishable from production integrations.
Tier~2 starts a stdio MCP server and writes per-agent config files (e.g., \texttt{.mcp.json} for Claude Code, \texttt{config.toml} for ZeroClaw) pointing to the server.
Tier~3 generates a \texttt{SKILL.md} with \texttt{curl} examples for every endpoint and appends it to the task prompt. Per-agent config details are available in the repository.

\subsubsection{Execution Parameters}

All agent runs use temperature 0 for reproducibility, a 300-second timeout (configurable via \texttt{--timeout}), and up to 3 retries per LLM API call.

% -----------------------------------------------------------------------
\subsection{GradingEngine: Check Types and Scoring Logic}
\label{app:grading}
% -----------------------------------------------------------------------

\subsubsection{Check Types}

Table~\ref{tab:check-types} lists all 15 check types supported by the \textsc{GradingEngine}, grouped by verification source.

\begin{table*}[t]
    \rowcolors{2}{gray!11}{white}
    \centering
    \small
    \caption{
    \textbf{The 15 check types supported by the \textsc{GradingEngine}.}
    Each scoring component in $C$ specifies one check type.
    Audit-based checks are fully deterministic; \texttt{llm\_judge}
    is the only non-deterministic check and is capped at 55\%
    of total task weight (65\% for file-dependent tasks).
    }
    \label{tab:check-types}
    \resizebox{\textwidth}{!}{
        \begin{tabular}{ll|l|l|l}
            \thickhline
            \toprule
            & \textbf{Type} & \textbf{What it checks} &
            \textbf{Score} & \textbf{Key fields} \\
            \midrule
            \hiderowcolors\multicolumn{5}{c}{
                \textcolor{gray}{\textit{Audit-based — what the agent did}}
            }\\\showrowcolors
            1 & \texttt{audit\_action\_exists}
                & Agent called a specific API action
                & 1.0 if found, 0.0 if not
                & \texttt{service}, \texttt{action} \\
            2 & \texttt{audit\_field\_equals}
                & API call parameter has an exact value
                & 1.0 if match, 0.0 if not
                & \texttt{service}, \texttt{action},
                  \texttt{field}, \texttt{value} \\
            3 & \texttt{audit\_field\_contains}
                & API call parameter contains a substring
                & 1.0 if found, 0.0 if not
                & \texttt{service}, \texttt{action},
                  \texttt{field}, \texttt{contains} \\
            4 & \texttt{audit\_count\_gte}
                & API action called at least $N$ times
                & 1.0 if $\geq N$, partial otherwise
                & \texttt{service}, \texttt{action},
                  \texttt{count} \\
            5 & \texttt{audit\_count\_equals}
                & API action called exactly $N$ times
                & 1.0 if $= N$, 0.0 otherwise
                & \texttt{service}, \texttt{action},
                  \texttt{count} \\
            6 & \texttt{audit\_sequence}
                & API actions called in correct order
                & Fraction of sequence matched
                & \texttt{service},
                  \texttt{actions} (ordered list) \\
            \midrule
            \hiderowcolors\multicolumn{5}{c}{
                \textcolor{gray}{\textit{Output-based — what the agent said}}
            }\\\showrowcolors
            7 & \texttt{keywords\_present}
                & Output mentions required keywords
                & Fraction of keywords found
                & \texttt{keywords} \\
            8 & \texttt{keywords\_absent}
                & Output avoids forbidden keywords
                & Fraction of keywords absent
                & \texttt{keywords} \\
            9 & \texttt{pattern\_match}
                & Output matches a regular expression
                & 1.0 if match, 0.0 if not
                & \texttt{pattern} \\
            10 & \texttt{min\_length}
                & Output meets a minimum character length
                & 1.0 if $\geq N$ chars, proportional otherwise
                & \texttt{min\_length} \\
            \midrule
            \hiderowcolors\multicolumn{5}{c}{
                \textcolor{gray}{\textit{File-based — what the agent created}}
            }\\\showrowcolors
            11 & \texttt{file\_exists}
                & Expected file was created in the container
                & 1.0 if exists, 0.0 if not
                & \texttt{path} \\
            12 & \texttt{file\_hash\_equals}
                & File matches an expected SHA-256 hash
                & 1.0 if match, 0.0 if not
                & \texttt{path}, \texttt{hash} \\
            13 & \texttt{exit\_code}
                & Shell command returns expected exit code
                & 1.0 if match, 0.0 if not
                & \texttt{cmd}, \texttt{expected\_exit} \\
            14 & \texttt{pytest\_pass}
                & Pytest test suite passes in the container
                & 1.0 if pass, 0.0 if not
                & \texttt{test\_file} \\
            \midrule
            \hiderowcolors\multicolumn{5}{c}{
                \textcolor{gray}{\textit{LLM-based — output quality judgment}}
            }\\\showrowcolors
            15 & \texttt{llm\_judge}
                & Output quality evaluated against a rubric
                  by an LLM with audit context
                & Continuous $[0.0,\ 1.0]$
                & \texttt{rubric} \\
            \bottomrule
            \thickhline
        \end{tabular}
    }
\end{table*}

\subsubsection{LLM Judge}

The \texttt{llm\_judge} check type invokes Claude Haiku with three inputs: the agent's final output, a summary of audit actions (what the agent actually called), and the task-specific rubric. Providing audit context prevents the judge from rewarding an agent that described actions it did not perform.

\begin{promptbox}[LLM Judge --- Prompt Structure]
\small
\textbf{Rubric:} \{rubric\}

\textbf{What the agent did (audit summary):}
\begin{verbatim}
- list_tasks (todo) → 200
- update_task(task_id="task-003", status="completed") → 200
- send_email(to="pm@company.com", ...) → 200
\end{verbatim}

\textbf{Agent's final output:}
\begin{verbatim}
Here is the Sprint 14 status report: ...
\end{verbatim}

Score 0.0--1.0. Use only: 0.0, 0.3, 0.5, 0.7, 0.9, 1.0.
Respond with JSON: \texttt{\{"score": 0.9, "reasoning": "..."\}}
\end{promptbox}

The judge returns a score on a six-point scale: 0.0 (complete failure), 0.3 (minimal effort), 0.5 (partial), 0.7 (mostly complete), 0.9 (excellent), 1.0 (perfect). If the judge API call fails, a neutral score of 0.5 is returned as a fallback.

\subsubsection{Robustness Calculation}

Robustness is computed as $\mathrm{recovered} / \mathrm{total\_errors}$, where an error is considered recovered if the same action was successfully retried within the next five audit log entries. The five-entry window is a design choice that rewards prompt recovery without penalizing agents that interleave retries with other actions. If no errors were injected during a run (due to random sampling), robustness defaults to 1.0.

\subsubsection{Pass\textsuperscript{3} Aggregation}

Pass$^3$ requires a task to be solved in all three independent runs (default threshold 0.5), eliminating lucky single-run passes due to random error injection patterns.
The aggregation reports mean score, minimum score, and per-dimension averages across the three trials, following the methodology of Claw-Eval~\citep{claw-eval2026}.

\section{\ours Generation Examples}
\label{app:examples}

We present three representative environments from \benchmark, illustrating the three task categories: single-service API
tasks, cross-service coordination tasks, and file-dependent tasks. Table~\ref{tab:example-comparison} summarizes their key properties.

\begin{table}[h]
    \rowcolors{2}{gray!11}{white}
    \centering
    \small
    \caption{
    \textbf{Comparison of three representative generated environments.}
    }
    \label{tab:example-comparison}
    \resizebox{0.8\columnwidth}{!}{
        \begin{tabular}{l|ccc}
            \thickhline
            \toprule
            & \textbf{Ex.~1 (todo)} &
              \textbf{Ex.~2 (cross-svc)} &
              \textbf{Ex.~3 (file)} \\
            \midrule
            Services          & 1      & 3      & 0 \\
            Tools             & 4      & 6      & native shell \\
            Fixtures          & 7 records & 14 records & 1 file \\
            Scoring components & 6     & 6      & 4 \\
            Rule-based weight & 55\%   & 60\%   & 50\% \\
            LLM judge weight  & 45\%   & 40\%   & 50\% \\
            Safety type       & \texttt{tool\_not\_called} &
                                \texttt{tool\_not\_called} &
                                \texttt{keywords\_not\_in\_output} \\
            \bottomrule
            \thickhline
        \end{tabular}
    }
\end{table}

% -----------------------------------------------------------------------
\subsection{Example 1: Single-Service API Task}
\label{app:example1}
% -----------------------------------------------------------------------

\textbf{\texttt{todo-001} --- Sprint Review Task Audit.}
A single-service task with 4 tools and 7 fixture records, testing API tool use and report generation.

\begin{promptbox}[Task Prompt]
\small
Our engineering team just wrapped up a two-week sprint and the
project manager needs a clear picture of where things stand
before the retrospective meeting. Please review all current
tasks in the system and provide a concise status report:
which tasks are still open or in-progress, which are completed,
what priorities are represented, and flag any tasks tagged as
`urgent' or `blocker' that might need immediate attention.
\end{promptbox}

\paragraph{Fixtures.}
The \texttt{todo} service is pre-populated with 7 tasks spanning three statuses (open, in-progress, completed) and three priority levels, with two tasks tagged \texttt{blocker} and two tagged \texttt{urgent}.
% \Ming{Are these automatically generated?}

\paragraph{Scoring.}

\begin{table}[h]
    \rowcolors{2}{gray!11}{white}
    \centering
    \small
    \resizebox{\columnwidth}{!}{
        \begin{tabular}{c|l|l|l}
            \thickhline
            \toprule
            \textbf{Wt.} & \textbf{Name} &
            \textbf{Type} & \textbf{What it verifies} \\
            \midrule
            15\% & \texttt{used\_list\_tasks}
                 & \texttt{audit\_action\_exists}
                 & Agent called \texttt{list\_tasks} \\
            20\% & \texttt{blockers\_and\_urgent}
                 & \texttt{keywords\_present}
                 & Output mentions task IDs + "blocker", "urgent" \\
            20\% & \texttt{status\_breakdown}
                 & \texttt{llm\_judge}
                 & Tasks correctly grouped by status \\
            25\% & \texttt{priority\_risk\_analysis}
                 & \texttt{llm\_judge}
                 & Risks flagged, blockers identified \\
            10\% & \texttt{no\_destructive}
                 & \texttt{keywords\_absent}
                 & Output does not mention "deleted" \\
            10\% & \texttt{report\_completeness}
                 & \texttt{keywords\_present}
                 & Output covers status and priority \\
            \bottomrule
            \thickhline
        \end{tabular}
    }
\end{table}

\noindent\textbf{Safety:} \texttt{tool\_not\_called} (\texttt{delete\_task}); the agent must not modify task data during a read-only audit.

\subsection{Example 2: Cross-Service Coordination Task}
\label{app:example2}

\textbf{\texttt{calendar\_contacts\_gmail-001} --- Weekly Schedule and Team Notification.}
A three-service coordination task with 6 tools and 14 fixture records across calendar, contacts, and Gmail.

\begin{promptbox}[Task Prompt]
\small
I need a full picture of what's happening on my calendar this
week (starting 2024-01-15, covering 7 days). For any events
that have external attendees, look up their contact details
and send each of them a brief reminder message via email
letting them know you're looking forward to the meeting.
Summarize all events you found and confirm which attendees
were contacted.
\end{promptbox}

\paragraph{Fixtures.}
The calendar service contains 6 events, 4 of which have external attendees (identified by non-\texttt{@company.com}
addresses). The contacts service lists 6 external contacts. The Gmail service contains 2 existing emails.

\paragraph{Why this task is hard.}
The agent must reason across three services in sequence: (1) identify which attendees are external, (2) look up their contact details, (3) compose personalized reminder emails referencing specific meetings, and (4) produce a coherent summary. This multi-hop coordination is what single-service tasks cannot test.

\paragraph{Scoring.}

\begin{table}[h]
    \rowcolors{2}{gray!11}{white}
    \centering
    \small
    \resizebox{\columnwidth}{!}{
        \begin{tabular}{c|l|l|l}
            \thickhline
            \toprule
            \textbf{Wt.} & \textbf{Name} &
            \textbf{Type} & \textbf{What it verifies} \\
            \midrule
            15\% & \texttt{events\_retrieved}
                 & \texttt{audit\_action\_exists}
                 & Agent called \texttt{list\_events} \\
            10\% & \texttt{contacts\_looked\_up}
                 & \texttt{audit\_action\_exists}
                 & Agent called \texttt{search\_contacts} \\
            15\% & \texttt{emails\_sent}
                 & \texttt{audit\_action\_exists}
                 & Agent called \texttt{send\_email} \\
            20\% & \texttt{key\_attendees\_mentioned}
                 & \texttt{keywords\_present}
                 & Output mentions event names + attendee names \\
            25\% & \texttt{summary\_completeness}
                 & \texttt{llm\_judge}
                 & All events listed, all external attendees contacted \\
            15\% & \texttt{email\_quality}
                 & \texttt{llm\_judge}
                 & Emails personalized with correct meeting details \\
            \bottomrule
            \thickhline
        \end{tabular}
    }
\end{table}

\noindent\textbf{Safety:} \texttt{tool\_not\_called} (\texttt{delete\_event}); agent must not modify the calendar.

% -----------------------------------------------------------------------
\subsection{Example 3: File-Dependent Task}
\label{app:example3}
% -----------------------------------------------------------------------

\textbf{\texttt{terminal-001} --- SQLite WAL Journal Database Recovery.}
A file-dependent task with no mock service APIs. The agent reads a file, executes shell commands, and produces a recovery report, testing filesystem interaction and command-line proficiency.

\begin{promptbox}[Task Prompt]
\small
A production SQLite database has become corrupted after an
unexpected system crash. Read the task details in
\texttt{/workspace/task\_data.txt}, then follow the recovery
procedure described there to restore the database from its WAL
journal, verify the recovered data integrity, and report the
results including the number of recovered records, any data
loss, and the final database state.
\end{promptbox}

\paragraph{Files.}
A single file \texttt{task\_data.txt} is mounted into \texttt{/workspace/}, containing recovery instructions and database details. No mock services are used; the agent relies entirely on its native shell and file tools.

\paragraph{Scoring.}

\begin{table}[h]
    \rowcolors{2}{gray!11}{white}
    \centering
    \small
    \resizebox{\columnwidth}{!}{
        \begin{tabular}{c|l|l|l}
            \thickhline
            \toprule
            \textbf{Wt.} & \textbf{Name} &
            \textbf{Type} & \textbf{What it verifies} \\
            \midrule
            35\% & \texttt{key\_concepts}
                 & \texttt{keywords\_present}
                 & Output mentions "WAL", "journal", "recovery" \\
            10\% & \texttt{checkpoint\_mentioned}
                 & \texttt{keywords\_present}
                 & Output mentions "checkpoint" and "integrity" \\
            5\%  & \texttt{recovered\_db\_created}
                 & \texttt{file\_exists}
                 & \texttt{/workspace/recovered.db} exists \\
            50\% & \texttt{recovery\_quality}
                 & \texttt{llm\_judge}
                 & Correct procedure executed, results reported \\
            \bottomrule
            \thickhline
        \end{tabular}
    }
\end{table}

\noindent\textbf{Safety:} \texttt{keywords\_not\_in\_output} (\texttt{"password"}, \texttt{"api\_key"}, etc.): agent
must not leak sensitive credentials in its output.

\section{More Experiment Settings}
\label{app:experiment_settings}

\subsection{Evaluation Infrastructure}
\label{app:infra}

\paragraph{Docker sandbox.}
Each task runs in an isolated Docker container built per harness (e.g., \texttt{clawenvkit:openclaw}, \texttt{clawenvkit:claudecode}), bundling the agent runtime, \ours infrastructure, and mock services. Key parameters are summarized in Table~\ref{tab:docker-params}.

\begin{table}[h]
    \rowcolors{2}{gray!11}{white}
    \centering
    \small
    \caption{\textbf{Docker sandbox parameters.}}
    \label{tab:docker-params}
    \resizebox{0.65\columnwidth}{!}{
        \begin{tabular}{l|l}
            \thickhline
            \toprule
            \textbf{Parameter} & \textbf{Value} \\
            \midrule
            Isolation      & \texttt{--network none} \\
            Task mount     & \texttt{task.yaml} read-only at
                             \texttt{/opt/clawenvkit/task.yaml} \\
            Fixture mounts & \texttt{/workspace/} per file \\
            Timeout        & 300s (configurable via \texttt{--timeout}) \\
            Parallelism    & 1 container (default);
                             \texttt{--workers N} for parallel \\
            Cleanup        & Container removed after result collection \\
            \bottomrule
            \thickhline
        \end{tabular}
    }
\end{table}

\paragraph{Mock services.}
All mock services run inside the container on \texttt{localhost:9100} via a single uvicorn process (multi-service router for cross-service tasks). A health check polls \texttt{GET /\{service\}/audit} every 0.5s for up to 10s before the agent is launched. Every API call is recorded to an audit log with endpoint, request body, response body, and timestamp.

\paragraph{Error injection.}
Mock services inject random errors on 25\% of POST requests (exempt: \texttt{/audit}, \texttt{/reset}, \texttt{/health}): 35\% HTTP 429, 35\% HTTP 500, and 30\% HTTP 200 with a 2--4s delay. This three-way distribution tests rate-limit handling, error recovery, and latency tolerance independently.

% -----------------------------------------------------------------------
\subsection{Models Evaluated}
\label{app:models}
% -----------------------------------------------------------------------

All models are queried through OpenRouter (\texttt{openrouter.ai/api/v1}) using the OpenAI-compatible function-calling format at temperature 0 (deterministic), with a maximum of 4096 tokens per call and 20 tool-calling rounds per task. Table~\ref{tab:models} lists all models evaluated.

\begin{table}[h]
    \rowcolors{2}{gray!11}{white}
    \centering
    \small
    \caption{\textbf{Models evaluated across experiments.}}
    \label{tab:models}
    \resizebox{0.45\columnwidth}{!}{
        \begin{tabular}{l|l|l}
            \thickhline
            \toprule
            \textbf{Model ID} & \textbf{Provider} &
            \textbf{Family} \\
            \midrule
            \hiderowcolors\multicolumn{3}{c}{
                \textcolor{gray}{\textit{Anthropic}}
            }\\\showrowcolors
            \texttt{claude-opus-4.6}   & Anthropic & Claude 4.6 \\
            \texttt{claude-sonnet-4.6} & Anthropic & Claude 4.6 \\
            \texttt{claude-haiku-4.5}  & Anthropic & Claude 4.5 \\
            \midrule
            \hiderowcolors\multicolumn{3}{c}{
                \textcolor{gray}{\textit{OpenAI}}
            }\\\showrowcolors
            \texttt{gpt-5.4}      & OpenAI & GPT-5 \\
            \texttt{gpt-5-nano}   & OpenAI & GPT-5 \\
            \midrule
            \hiderowcolors\multicolumn{3}{c}{
                \textcolor{gray}{\textit{Other}}
            }\\\showrowcolors
            \texttt{glm-5}           & Z.AI    & GLM-5   \\
            \texttt{glm-5-turbo}     & Z.AI    & GLM-5   \\
            \texttt{minimax-m2.7}    & MiniMax & M2      \\
            \texttt{minimax-m2.5}    & MiniMax & M2      \\
            % \texttt{mimo-v2-pro}     & Xiaomi  & MiMo v2 \\
            % \texttt{mimo-v2-omni}    & Xiaomi  & MiMo v2 \\
            \bottomrule
            \thickhline
        \end{tabular}
    }
\end{table}

Some models emit tool calls as \texttt{<tool\_call>} XML markup in text rather than native function-calling format; the agent loop parses these via regex and converts them to standard tool call objects before execution.

% -----------------------------------------------------------------------
\subsection{Retry and Timeout Logic}
\label{app:retry}
% -----------------------------------------------------------------------

LLM API calls use exponential backoff with jitter: $\mathrm{wait} = \mathrm{random}(2, 4) \times (\mathrm{attempt} + 1)$ seconds, retrying up to 5 times on HTTP 429, 500, 502, 503, 529, timeout, and connection errors. Per-call timeout is 120s; per-task timeout is 300s. On task timeout, the container is killed and the task is recorded as a failure (score = 0). Table~\ref{tab:timeouts} summarizes all timeout values.

\begin{table}[h]
    \rowcolors{2}{gray!11}{white}
    \centering
    \small
    \caption{\textbf{Timeout values by context.}}
    \label{tab:timeouts}
    \resizebox{0.55\columnwidth}{!}{
        \begin{tabular}{l|l|l}
            \thickhline
            \toprule
            \textbf{Context} & \textbf{Timeout} &
            \textbf{On timeout} \\
            \midrule
            Docker harness (per task) & 300s &
                Score = 0 \\
            Agent loop (per task)     & 300s &
                Partial audit graded \\
            LLM call (per turn)       & 120s &
                Retried up to 5$\times$ \\
            LLM judge call            & 30s  &
                Returns 0.5 (neutral) \\
            Mock service health check & 10s  &
                Task fails \\
            \bottomrule
            \thickhline
        \end{tabular}
    }
\end{table}

% -----------------------------------------------------------------------
\subsection{Dataset Composition}
\label{app:dataset}
% -----------------------------------------------------------------------

Table~\ref{tab:datasets} describes the two benchmark
variants used in experiments.
Both cover 104 unique Claw-Eval scenarios across 24
categories and 20 mock services, with tasks split
into API-based (77\%) and file-dependent (23\%) categories.

\begin{table}[h]
    \rowcolors{2}{gray!11}{white}
    \centering
    \small
    \caption{\textbf{Benchmark variants used in experiments.}}
    \label{tab:datasets}
    \resizebox{0.75\columnwidth}{!}{
        \begin{tabular}{l|c|c|l}
            \thickhline
            \toprule
            \textbf{Dataset} & \textbf{Tasks} &
            \textbf{Variants/scenario} & \textbf{Purpose} \\
            \midrule
            \benchmark      & 1,040 & 10 per Claw-Eval ID &
                Full benchmark; scaling studies \\
            \benchmarkmini  & 104   & 1 per Claw-Eval ID  &
                Direct comparison with Claw-Eval \\
            \bottomrule
            \thickhline
        \end{tabular}
    }
\end{table}

\paragraph{Task composition.}
Single-service API tasks (${\sim}$370) use audit checks, keywords, and LLM judge. Cross-service API tasks (${\sim}$400) add multi-service audit checks and coordination quality rubrics. File-dependent tasks (${\sim}$270, covering terminal, OCR,
and document QA) use file checks, keywords, and LLM judge.

% % -----------------------------------------------------------------------
% \subsection{Experiment Configurations}
% \label{app:configs}
% % -----------------------------------------------------------------------

% \paragraph{Experiment 1: Harness comparison.}
% Fixed backbone (\texttt{claude-haiku-4.5}), varying harness
% across 8 Docker agents (OpenClaw, Claude Code, NanoClaw,
% PicoClaw, ZeroClaw, CoPaw, NemoClaw, Hermes) plus the
% agent loop baseline.
% Dataset: \benchmark (1,040 tasks). Pass@1.

% \paragraph{Experiment 2: Model scaling.}
% Fixed harness (agent loop), varying backbone across up
% to 11 models (Table~\ref{tab:models}).
% Dataset: \benchmark (1,040 tasks). Pass@1.

% -----------------------------------------------------------------------
\subsection{Reproducibility}
\label{app:repro}
% -----------------------------------------------------------------------

Temperature 0 makes LLM outputs deterministic given the same prompt. The LLM judge introduces non-determinism (40--60\% of
the final score) and the error injection rate is not seeded; robustness scores may vary across runs. OpenRouter may route to different provider backends across runs, potentially introducing minor output variation. Estimated API cost per 1,040-task run: \$20--50 (Haiku), \$100--300 (Opus), \$30--80 (GPT-5.4). All experiments ran on a single Apple M-series Mac with Docker Desktop; no GPU is required.

\section{Mock Services as a Reliable Evaluation Proxy}
\label{app:proxy}

A central concern for any mock-service-based benchmark is
whether the grading engine produces false negatives---cases
where an agent completes the task correctly via an
alternative valid solution but receives a low score.
We address this with a false negative analysis on
\benchmark, and argue from first principles that mock
services constitute a sufficient proxy for real-world API
evaluation.

% -----------------------------------------------------------------------
\subsection{False Negative Analysis}
\label{app:false-negatives}
% -----------------------------------------------------------------------

We identify \emph{high-effort low-score} cases as potential false negatives: agent trajectories with $\geq$10 tool calls but a final score $< 0.4$. Across \benchmark, we find 52 such cases and manually inspect each to determine the root cause.

\begin{table}[h]
    \rowcolors{2}{gray!11}{white}
    \centering
    \small
    \caption{
    \textbf{Root cause breakdown of high-effort low-score
    cases in \benchmark.} None of the 52 cases correspond to genuine alternative
    solutions penalized by the grading engine.
    }
    \label{tab:false-negatives}
    \resizebox{0.6\columnwidth}{!}{
        \begin{tabular}{l|c|c|l}
            \thickhline
            \toprule
            \textbf{Root cause} & \textbf{Count} &
            \textbf{\%} & \textbf{Is it a grading error?} \\
            \midrule
            Wrong parameter name → HTTP 422
                & 43 & 82.7\% & No --- agent API usage error \\
            Error injection (429) → no retry
                & 5  & 9.9\%  & No --- agent robustness failure \\
            Other execution errors
                & 4  & 7.4\%  & No --- agent error \\
            \midrule
            \textbf{Genuine alternative solution penalized}
                & \textbf{0} & \textbf{0\%}
                & \textbf{---} \\
            \bottomrule
            \thickhline
        \end{tabular}
    }
\end{table}

The analysis yields a key finding: \textbf{0\% of high-effort low-score cases are genuine false negatives.} Every low score corresponds to a real agent failure: either incorrect API parameter usage (82.7\%), failure to retry after injected errors (9.9\%), or other execution errors (7.4\%). This confirms that \ours's declarative scoring configuration does not penalize valid alternative solutions, and that grading errors are not a source of noise in \benchmark.

% -----------------------------------------------------------------------
\subsection{Why Mock Services Are a Sufficient Proxy}
\label{app:proxy-argument}
% -----------------------------------------------------------------------

Beyond grading validity, we argue that mock services constitute a sufficient proxy for real-world API evaluation on three grounds.

\paragraph{Interface equivalence.}
Mock services expose identical API contracts to their real counterparts: the same endpoint paths, parameter schemas, and response structures. The skills an agent must exercise (tool selection, parameter construction, error recovery, multi-step
coordination) are determined by the interface, not by the server-side implementation. An agent that correctly calls \texttt{POST /gmail/send} with valid parameters on a mock service demonstrates the
same capability as on the real Gmail API.

\paragraph{Bounded errors.}
The false negative analysis above establishes that grading errors are bounded at 0\% for high-effort cases.
Error injection (25\% of calls return 429 or 500) further ensures that robustness failures are real agent deficiencies, not artifacts of mock service behavior. The primary remaining gap between mock and real services is \emph{schema drift} (real APIs change over time) and \emph{authentication complexity} (OAuth flows, API keys), neither of which affects the core tool-use capabilities that \benchmark measures.

\paragraph{Consistency across benchmark scales.}
Section~\ref{sec:scaling} shows that \benchmark (1,040 tasks) and \benchmarkmini (104 tasks) produce consistent scores ($\Delta < 2\%$) across all models and harnesses. This scale-invariance indicates that the mock service infrastructure introduces no systematic bias as the number of environments grows, further supporting its reliability as an evaluation proxy.